\documentclass[11pt]{article}

\PassOptionsToPackage{table}{xcolor}
\usepackage[preprint]{acl}

\usepackage{times}
\usepackage{latexsym}

\usepackage[T1]{fontenc}

\usepackage[utf8]{inputenc}

\usepackage{microtype}

\usepackage{inconsolata}

\usepackage{graphicx}
\usepackage{subcaption}

\usepackage{amsmath}
\usepackage{amssymb}
\usepackage{booktabs}
\usepackage{multirow}
\usepackage{enumitem}
\usepackage{colortbl}
\usepackage{url}
\usepackage{algorithm}
\usepackage{algorithmic}

\newcommand{\model}{MAGIC$^3$}
\newcommand{\fullmodel}{Modal-Adversarial Gated Interaction and Consistency-Centric Classifier}

\definecolor{magicrow}{gray}{0.92}
\newcommand{\best}[1]{\textbf{#1}}
\newcommand{\second}[1]{\underline{#1}}

\definecolor{revision}{RGB}{0,0,0}

\title{Exposing Cross-Modal Consistency\\ for Fake News Detection in Short-Form Videos}

\author{
  \textbf{Chong Tian\textsuperscript{1}},
  \textbf{Yu Wang\textsuperscript{1}},
  \textbf{Chenxu Yang\textsuperscript{2}},
  \textbf{Junyi Guan\textsuperscript{1}},
\\
  \textbf{Zheng Lin\textsuperscript{2}},
  \textbf{Yuhan Liu\textsuperscript{1}},
  \textbf{Xiuying Chen\textsuperscript{1}\thanks{Corresponding authors.}},
  \textbf{Qirong Ho\textsuperscript{1}\footnotemark[1]}
\\
\\
  \textsuperscript{1}Mohamed bin Zayed University of Artificial Intelligence (MBZUAI)
\\
  \textsuperscript{2}Institute of Information Engineering, Chinese Academy of Sciences (CAS)
\\
  \small
    \texttt{\{chong.tian, xiuying.chen, qirong.ho\}@mbzuai.ac.ae}
}

\begin{document}
\maketitle

\begin{abstract}
Short-form video platforms are major channels for news but also fertile ground for multimodal misinformation where each modality appears plausible alone yet cross-modal relationships are subtly inconsistent, like mismatched visuals and captions.
On two benchmark datasets, FakeSV (Chinese) and FakeTT (English),  we observe a clear asymmetry: real videos exhibit high text–visual but moderate text–audio consistency, while fake videos show the opposite pattern. 
Moreover, a single global consistency score forms an interpretable axis along which fake probability and prediction errors vary smoothly.
Motivated by these observations, we present \textbf{\model{}} (\fullmodel{}), a detector that explicitly models and exposes cross-tri-modal consistency signals at multiple granularities.
\model{} combines explicit pairwise and global consistency modeling with token- and frame-level consistency signals derived from cross-modal attention, incorporates multi-style LLM rewrites to obtain style-robust text representations, and employs an uncertainty-aware classifier for selective VLM routing.
Using pre-extracted features, \model{} consistently outperforms the strongest non-VLM baselines on FakeSV and FakeTT.
While matching VLM-level accuracy, the two-stage system achieves 18–27× higher throughput and 93\% VRAM savings, offering a \emph{strong cost–performance tradeoff}.

\end{abstract}

\section{Introduction}

\begin{figure}[t]
  \centering
  \includegraphics[width=\columnwidth]{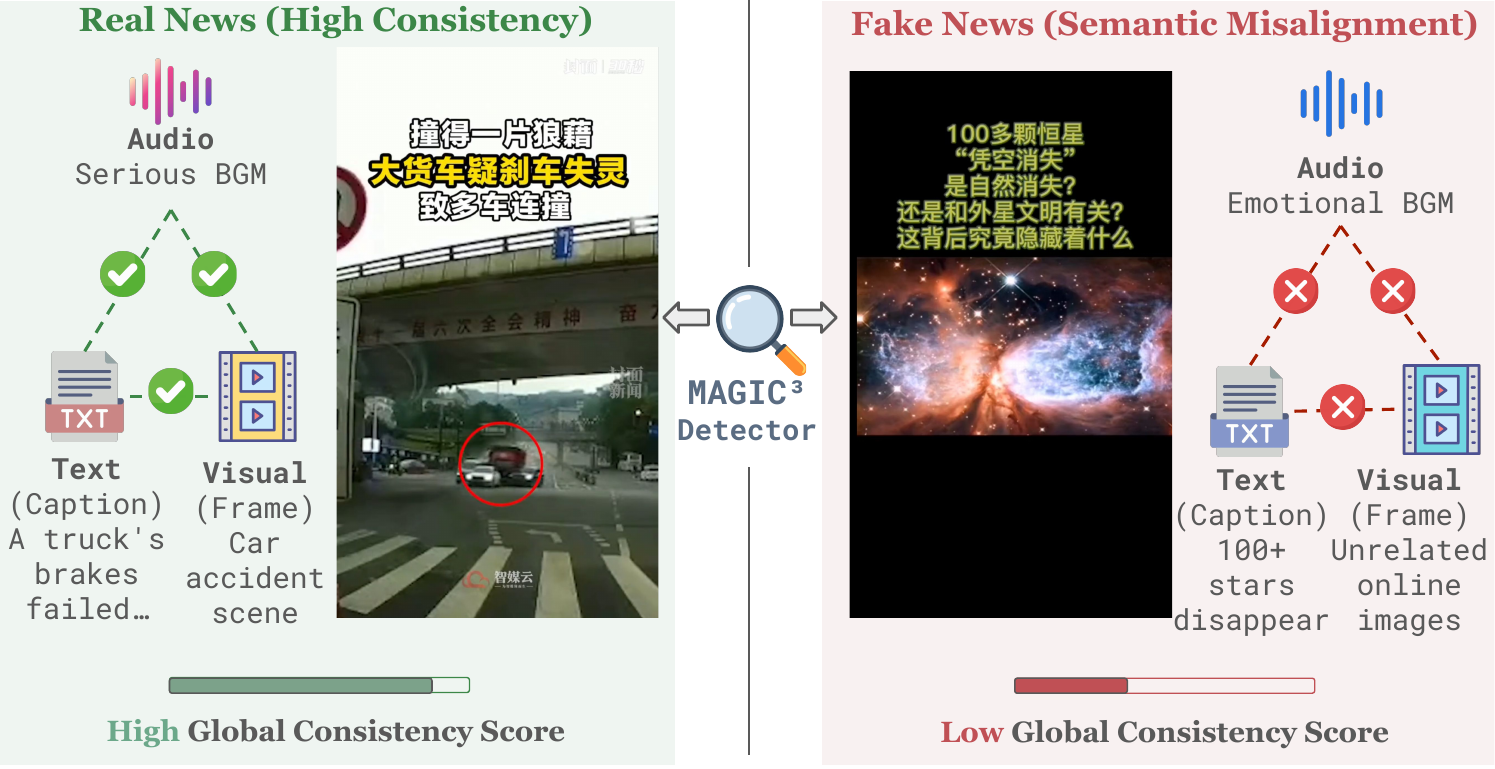}
  \caption{\textbf{Illustration of cross-modal consistency patterns.} In real news short videos, text, visuals, and audio are contextually aligned (Consistent). In fake news, a ``semantic gap'' often exists between the sensational claims (text/audio) and the actual visual content. \model{} acts as a consistency lens to quantify these multimodal relationships for fake news detection.}
  \label{fig:motivation}
\end{figure}
Short-form video platforms such as TikTok, Douyin, YouTube Shorts, and Kuaishou have reshaped how people consume news and public information~\citep{bu2023combating}. Given the sheer volume of uploads, detection systems must be not only accurate but also computationally efficient.  
Rich multimodal signals and highly optimized recommendation systems allow fake news \emph{short videos} to attract massive attention before professional fact-checking becomes available, amplifying societal risks~\citep{vosoughi2018spread,zhou2020survey}.  
Compared to text-only fake news~\citep{devlin2019bert,shu2019beyond,tian2025symbolic}, short news videos are harder to analyze: manipulations may occur in any modality and at any granularity, and successful fakes rarely contain glaring artifacts in a single stream.  
Instead, they often exploit \emph{cross-modal inconsistencies}, where each modality looks plausible alone but the joint message is misleading---as illustrated in Figure~\ref{fig:motivation}.


\paragraph{A consistency-centric view.}
We adopt a task-oriented taxonomy of cross-modal inconsistency in short-form news videos, covering entity/context, event semantics, affect, temporal audio--visual misalignment, and style/pragmatics, with detailed definitions and examples provided in Appendix~\ref{sec:taxonomy-appendix}.
Throughout this paper, we use ``consistency'' to denote \emph{learned, task-oriented compatibility signals in feature space}, rather than factual correctness with respect to external world knowledge.  
Instead of designing yet another heavy backbone, we build a lightweight, interpretable \emph{consistency lens} that sits on top of any multimodal feature extractor and is tailored to fake news short videos.

\paragraph{Key findings.}
Grounded in this consistency-centric view, our analysis reveals four key insights:
(i) \textit{Asymmetric consistency}: real videos show high text--visual but moderate text--audio consistency, while fake videos flip this pattern;
(ii) \textit{Interpretable axis}: global consistency strongly correlates with prediction difficulty, with errors clustering at intermediate values;
(iii) \textit{Efficient routing}: combining consistency with uncertainty enables routing only 25\% of samples to heavyweight VLMs while \emph{surpassing} VLM-only accuracy;
(iv) \textit{Style robustness}: multi-style LLM rewrites improve robustness, with fake videos showing higher consistency variance.

\paragraph{Our method.}
We introduce \textbf{\model{}}, a feature-level detector designed around explicit consistency modeling.  
Given frozen text, visual, and audio features (and a few LLM-rewritten texts), \model{} (i) computes pairwise and global consistency scores through the Cross-Modal Consistency Gate (CMCG), (ii) derives token- and frame-level \emph{consistency fields} from cross-modal attention, (iii) builds style-robust text representations in an Adversarial-Aware Rewrite Fusion (AARF) module, (iv) performs lightweight hierarchical multimodal fusion with a temporal audio--visual inconsistency score (TCMI), and (v) exposes calibrated uncertainty estimates that drive a two-stage pipeline where expensive VLM-based detectors are invoked only when needed. 

\paragraph{Contributions.}
We make three contributions:
(i) To our knowledge, this is the first work on fake news short videos that makes tri-modal (text--visual--audio) consistency an \emph{explicit}, multi-granularity output: pairwise and global scalar scores, plus token- and frame-level consistency fields that reveal text--visual vs.\ text--audio asymmetries and an interpretable consistency--difficulty axis.
(ii) \model{}, a detector that builds on these explicit consistency signals together with calibrated uncertainty estimates, providing an interpretable ``consistency lens'' for short-form news videos.
(iii) A consistency- and uncertainty-driven two-stage VLM routing strategy that reaches state-of-the-art accuracy on FakeSV and FakeTT while offering 18–27× higher throughput, offering substantially better cost--performance tradeoff.

\section{Related Work}

\paragraph{Multimodal fake news on image--text pairs.}
Prior work extends fake news detection to image–text posts by jointly modeling captions and images~\citep{zhou2020safe,muller2020multimodal,li2025survey}.
Event-adversarial methods such as EANN~\citep{wang2018eann} learn event-invariant multimodal features to improve generalization, while similarity-aware approaches like SAFE~\citep{zhou2020safe} encode image–text agreement or mismatch to flag manipulations.
These methods motivate our focus on explicit cross-modal consistency, but they mainly target article- or post-level content rather than short videos with audio.

\paragraph{Short news video detection.}
Existing fake news detectors on short videos fall into three families.
Feature-based multimodal models use strong single-modality encoders with late fusion like FakeSV~\citealp{qi2023fakesv}, but weakly exploit cross-modal interactions.
Deep fusion architectures such as TikTec, FANVM, SVFEND, FakingRecipe, and KDSGAT-FNVD~\citep{shang2021multimodal,choi2021using,qi2023fakesv,bu2024fakingrecipe,huang2025knowledge} learn powerful joint representations and \emph{implicitly} reason about consistency, but do not expose where inconsistencies arise.
More recent systems, like CA-FVD~\citep{wang2025consistency}, GLFE-SVFD~\citep{wang2025global}, and SCMG-FND~\citep{yang2025fake}, leverage consistency-aware training, global--local feature enhancement, or LLM-based semantic credibility scores to better capture multimodal relationships, yet their consistency cues remain coarse video-level signals or internal features.
In contrast, we focus on short news videos with three modalities and turn multi-granularity consistency---pairwise/global scores, token-/frame-level consistency fields, and temporal audio--visual inconsistency---into explicit outputs that also drive an uncertainty-aware VLM routing.


\begin{figure*}[t]
  \centering
  \includegraphics[width=0.75\textwidth]{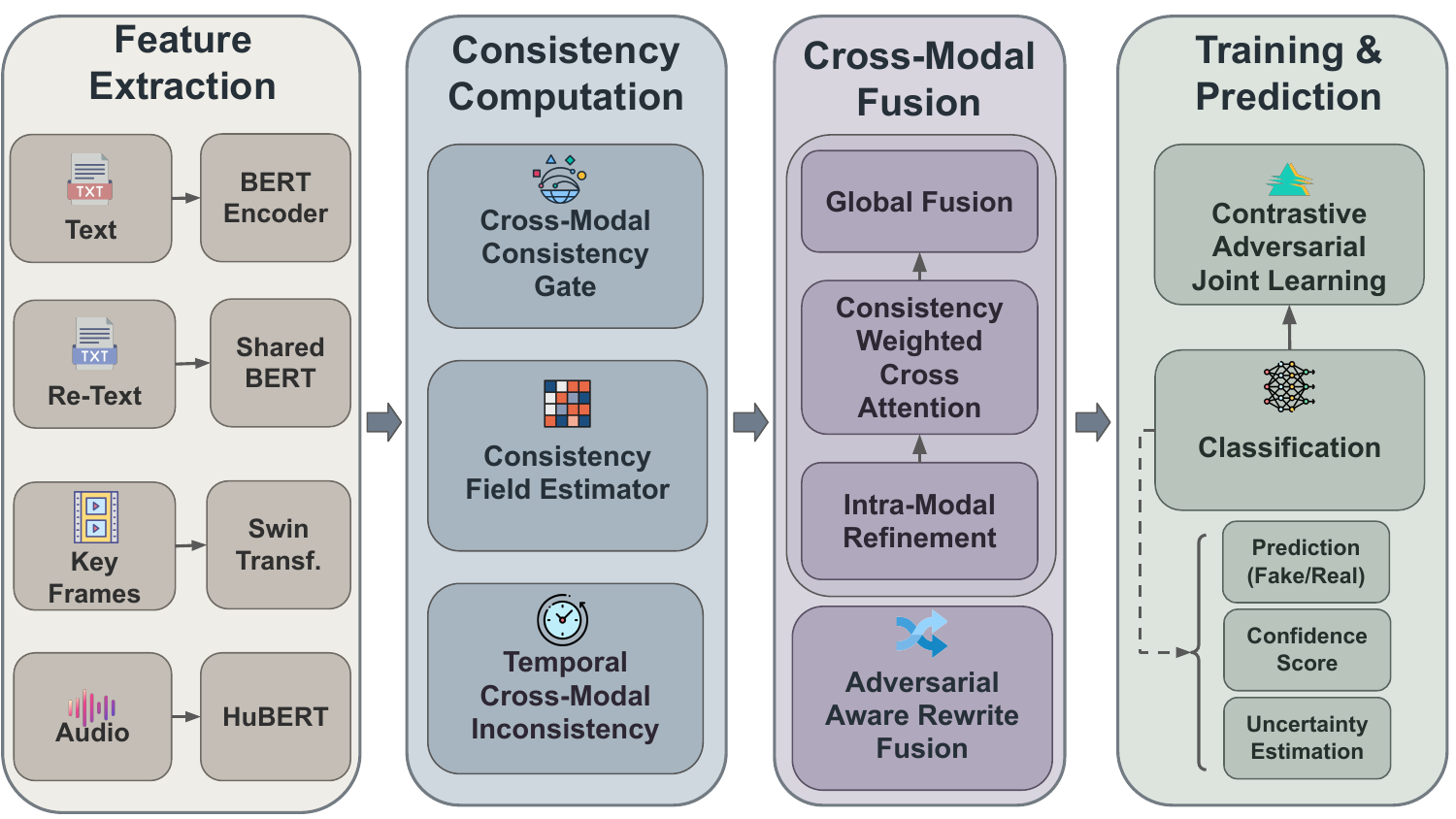}
  \caption{\model{} Overview. Frozen encoders provide text, visual, audio, and rewrite features. The Cross-Modal Consistency Gate outputs pairwise and global consistency scores; Consistency Field Estimator converts cross-modal attention into token- and frame-level consistency fields; Temporal Cross-Modal Inconsistency computes a temporal inconsistency score. Adversarial Aware Rewrite Fusion fuses original text with LLM rewrites into a style-robust representation; Hierarchical Multimodal Transformer performs hierarchical multimodal fusion; and the classifier outputs fake probability and uncertainty. The framework is trained via Contrastive Adversarial Joint Learning.}
  \label{fig:architecture}
\end{figure*}

\section{Methodology}
\label{sec:methodology}

\subsection{Problem Formulation}
\label{sec:problem_formulation}
In this work, we focus on short-form news videos, as they constitute a dominant and practically important modality for online news consumption and multimodal misinformation.
Both benchmarks we study, FakeSV and FakeTT, are constructed specifically from short videos on real-world platforms and reflect how users actually encounter news in today's media ecosystem. Our modeling choices, findings, and efficiency claims should therefore be interpreted within this practically important but intentionally scoped domain. 
Given a short news video $\mathcal{V}$ with associated text $\mathcal{T}$ (titles, subtitles, and descriptions), the task is to predict whether it is fake ($y=1$) or real ($y=0$).

\subsection{Feature Extraction}
\label{subsec:feature_extraction}
Following prior work~\citep{qi2023fakesv,huang2025knowledge}, we employ fixed pretrained encoders to extract feature-level representations. 
Specifically, we use BERT~\citep{devlin2019bert} for encoding text titles and descriptions, Swin Transformer~\citep{liu2021swin} for encoding visual keyframes, and ExHuBERT~\citep{amiriparian2024exhubert} for audio emotion embeddings.
This decoupled design lets \model{} leverage state-of-the-art representations without the burden of end-to-end fine-tuning. 
To enhance robustness, we also generate three LLM rewrites of each text (neutral, formal, sensational) and encode them with the same text encoder. 

\subsection{Consistency Computation}
\label{subsec:consistency_computation}
This module computes multi-granularity consistency signals to expose cross-modal discrepancies at different abstraction levels.
We hypothesize that fake news videos exhibit inconsistencies in diverse ways: some are globally incoherent (e.g., unrelated subtitles and visuals), while others contain subtle, localized manipulations (e.g., a specific phrase contradicting a visual frame) or temporal artifacts.
To capture this spectrum, we design three complementary mechanisms: (1) a global gate for high-level filtering, (2) fine-grained consistency fields for localized grounding, and (3) temporal inconsistency scores to detect synchronization artifacts.
This multi-view approach ensures that \model{} remains sensitive to both cheapfakes and sophisticated manipulations.

\textbf{Cross-Modal Consistency Gate} (CMCG). 
As shown in Figure~\ref{fig:architecture}, this module pools modality features into vectors $\bar{\mathbf{h}}_{\mathrm{text}}$, $\bar{\mathbf{h}}_{\mathrm{vis}}$, and $\bar{\mathbf{h}}_{\mathrm{aud}}$. It then computes scalar consistency scores for each pair $(m_i,m_j)$ via a gated mechanism:
\begin{equation}
  c_{ij} = \sigma(\mathbf{w}_{ij}^{\top} [\bar{\mathbf{h}}_{m_i};\bar{\mathbf{h}}_{m_j}] + b_{ij}).
\end{equation}
These pairwise scores form a vector $\mathbf{c} = [c_{\mathrm{tv}}, c_{\mathrm{ta}}, c_{\mathrm{va}}]$, which provides a compact summary of inter-modal agreement. Finally, they are compressed into a single global consistency scalar:
\begin{equation}
  c_{\mathrm{global}} = \sigma(\mathrm{MLP}_{c}(\mathbf{c})).
\end{equation}
This global score $c_{\mathrm{global}}$ provides a high-level measure of the video's overall coherence and modulates cross-modal attention. Throughout, we use "consistency'' to denote learned, task-oriented compatibility proxies computed in frozen encoder feature spaces, keeping the consistency lens lightweight and decoupled from heavy backbones.

\textbf{Consistency Field Estimator } (CFE). 
It converts cross-modal attention matrices into per-token and per-frame ``consistency fields'', localizing \emph{where} a video is inconsistent.  
Given a cross-attention matrix $\mathbf{A}_{m_i \leftarrow m_j} \in \mathbb{R}^{L_{m_i} \times L_{m_j}}$ from the cross-modal transformer, we define:
\begin{equation}
  F_{ij}^{(m_i)}(t)
  =
  \max_{k}\, \mathbf{A}_{m_i \leftarrow m_j}(t,k),
  \quad t = 1,\dots,L_{m_i}.
\end{equation}
Intuitively, $F_{ij}^{(m_i)}(t)\in[0,1]$ reflects how strongly token $t$ in modality $m_i$ finds supporting evidence in modality $m_j$.  
Tokens that attend strongly and with high focus receive high scores; tokens that attend diffusely or weakly receive low scores.  
These fields support heatmap visualizations and localized analysis, and visualizations of some examples of FakeSV and FakeTT are shown in Figure~\ref{fig:cfe-fields} in Appendix~\ref{sec:addl-figures}.

\textbf{Temporal Cross-Modal Inconsistency }(TCMI).  
In parallel, we align audio and visual features along time to capture coarse mismatches. We resample both modalities to length $T'$ and compute per-timestep distances:
\begin{equation}
  d_t = \bigl\|
    \tilde{\mathbf{h}}_{\mathrm{vis},t}
    - \mathrm{Proj}_{\mathrm{av}}(\tilde{\mathbf{h}}_{\mathrm{aud},t})
  \bigr\|_2,
\end{equation}
which are summarized into $\mathbf{s}_{\mathrm{temp}} = [\mathrm{mean}(\mathbf{d}); \mathrm{var}(\mathbf{d}); \max(\mathbf{d})]$ and mapped to:
\begin{equation}
  c_{\mathrm{temp}} = 1 - \sigma\bigl(\mathrm{MLP}_{\mathrm{temp}}(\mathbf{s}_{\mathrm{temp}})\bigr).
\end{equation}
This captures coarse audio--visual mismatches inspired by deepfake detectors~\citep{gao2024temporal}.

\subsection{Cross-Modal Fusion}
\label{subsec:cross_modal_fusion}

\textbf{Adversarial-Aware Rewrite Fusion }(AARF).
The Multi-view Adversarial-Aware Rewrite Fusion module treats LLM rewrites as semantic-preserving but style-altering views (like in neutral, formal, and sensational styles).  
Let $\mathbf{h}_{\mathrm{orig}} = \mathrm{Pool}(\mathbf{H}_{\mathrm{text}})$ be the original text and $\mathbf{h}_{\mathrm{rew}}^{(v)}$ the $v$-th rewrite. We compute gated weights:
\begin{equation}
  q^{(v)} = \sigma\!\bigl(\mathrm{MLP}_{q}[\mathbf{h}_{\mathrm{orig}};\mathbf{h}_{\mathrm{rew}}^{(v)}]\bigr),
\end{equation}
where $q^{(v)}$ estimates rewrite quality. A gating MLP produces fusion logits from all views, and we enforce a minimum weight $\alpha_{\min}$ on the original text:
\begin{equation}
  \mathbf{h}_{\mathrm{text}}^{\mathrm{fuse}}
  =
  \alpha_{0}\,\mathbf{h}_{\mathrm{orig}}
  +
  \sum_{v=1}^{V} \alpha_{v}\,\mathrm{Proj}\bigl(\mathbf{h}_{\mathrm{rew}}^{(v)}\bigr),
\end{equation}
where $\alpha_0 \geq \alpha_{\min}$ and $\sum_v \alpha_v = 1$, and $\mathrm{Proj}(\cdot)$ is a learned linear projection that aligns the rewrite embeddings with the original text embedding dimension.
Contrastive losses pull together representations of rewrites from the same video and push apart those from different videos, encouraging style-invariant embeddings.

\textbf{Hierarchical Multimodal Transformer }(HMT).  
This component operates in three stages.
(i) \emph{Intra-modal refinement}: for each modality $m \in \{\text{text}, \text{visual}, \text{audio}\}$, we apply a modality-specific Transformer encoder with self-attention,
$\mathbf{H}_{m}^{(A)} = \mathrm{TransEnc}_{m}(\mathbf{H}_{m})$,
to refine token or frame sequences within that modality.
(ii) \emph{Consistency-weighted cross-attention}: we then add cross-attention blocks where queries from one modality attend to keys/values from another modality, and aggregate the resulting messages using weights derived from the pairwise consistency vector $\mathbf{c}$. The cross-attention matrices in this stage are also used by CFE to construct token- and frame-level consistency fields.
(iii) \emph{Global aggregation}: finally, we pool the refined representations from all modalities (and the fused text) and feed them into a small global fusion layer to obtain a single video-level representation $\mathbf{h}_{\mathrm{global}}$ that is passed to the classifier.

\subsection{Training and Prediction}
\label{subsec:training_prediction}

We train \model{} using a Contrastive--Adversarial Joint Learning (CAJL) objective to jointly optimize classification accuracy and the reliability of consistency and uncertainty signals for prediction and routing.
The primary supervised loss is standard binary cross-entropy:
\begin{equation}
  \mathcal{L}_{\mathrm{ce}}
  =
  -\frac{1}{N}
  \sum_{i}
  \Bigl(
    y_i \log \hat{y}_i
    + (1-y_i)\log (1-\hat{y}_i)
  \Bigr).
\end{equation}
To improve robustness and alignment, we employ InfoNCE-style contrastive losses for intra- and cross-modal alignment ($\mathcal{L}_{\mathrm{intra}}, \mathcal{L}_{\mathrm{cross}}$). For a batch of $N$ pairs $(\mathbf{u}_i, \mathbf{v}_i)$, the loss is:
\begin{equation}
  \mathcal{L}_{\mathrm{NCE}}
  =
  -\frac{1}{N}
  \sum_{i}
  \log
  \frac{\exp(\mathrm{sim}(\mathbf{u}_i,\mathbf{v}_i) / \tau)}
  {\sum_j \exp(\mathrm{sim}(\mathbf{u}_i,\mathbf{v}_j)/\tau)}.
\end{equation}
To improve prediction robustness, we also introduce random noise perturbations to the hidden representations during training and minimize the KL-divergence between predictions on clean and perturbed features, formulated as an adversarial consistency regularizer $\mathcal{L}_{\mathrm{adv}}$. 
The final composite objective combines these terms with a consistency regularizer $\mathcal{L}_{\mathrm{reg}}$ that aligns global and local scores:
\begin{align}
  \mathcal{L}
  &=
  \mathcal{L}_{\mathrm{ce}}
  +
  \lambda_{\mathrm{intra}} \mathcal{L}_{\mathrm{intra}}
  +
  \lambda_{\mathrm{cross}} \mathcal{L}_{\mathrm{cross}} \nonumber \\
  &+
  \lambda_{\mathrm{adv}} \mathcal{L}_{\mathrm{adv}}
  +
  \lambda_{\mathrm{reg}} \mathcal{L}_{\mathrm{reg}},
\end{align}
where $\lambda$ are trade-off hyperparameters. More details on all terms are provided in Appendix~\ref{sec:method-details}.

\textbf{Classifier.}
The final classifier produces three distinct signals:
(1) Prediction: The binary classification probability $\hat{y}_{fake}$ indicating the likelihood of the video being fake.
(2) Confidence Score: Derived from the softmax output as $\text{conf} = \max(\hat{y}, 1-\hat{y})$, representing the model's confidence in its chosen class.
(3) Uncertainty Estimation: A composite metric combining predictive entropy $u_{\mathrm{ent}} = -\sum p \log p$ and the learned auxiliary scalar $\hat{u}$ (trained to regress the error margin). 
These signals provide a robust basis for the subsequent routing decision, allowing the system to distinguish between confident errors and genuine ambiguity. 

\textbf{Two-stage Routing.}
During inference, we employ a routing function $\mathcal{R}(x)$ to decide whether to trust \model{}'s prediction or escalate to a heavyweight VLM. Unlike simple confidence thresholding, we leverage our multi-dimensional outputs to define effective routing strategies (detailed in Appendix~\ref{subsec:finding3-routing}):
(1) \textit{Uncertainty-based}: Routes samples with high predictive entropy ($u_{\mathrm{ent}} > \tau$), prioritizing cases where the model is information-theoretically uncertain.
(2) \textit{Difficulty-based}: Uses a composite difficulty score $d = \text{Norm}(u_{\mathrm{ent}}) + (1 - c_{\mathrm{global}}) + (1-\text{conf})$ to route samples that are simultaneously uncertain, inconsistent, and low-confidence.
By filtering out the $\sim$75\% of "easy" cases that \model{} handles correctly, we achieve VLM-level accuracy at a fraction of the cost.

\section{Experiments}

\subsection{Experimental Setup}

We evaluate \model{} on the FakeSV and FakeTT benchmarks for short video fake news detection~\citep{qi2023fakesv,huang2025knowledge}.  
We follow the chronological split protocol of~\citet{huang2025knowledge}, using 70\%/15\%/15\% of videos for training, validation, and testing.  
Dataset statistics are summarized in Table~\ref{tab:datasets} and detailed further in Appendix~\ref{sec:datasets}.

Text, visual, and audio features are extracted as described in Section~\ref{subsec:feature_extraction} using fixed pretrained encoders.  
All trainable parameters reside in the \model{} detector head (CMCG, CFE, AARF, HMT+TCMI, and the classifier).  
We use AdamW with learning rate $5\times 10^{-5}$, batch size 32, cosine decay with 10\% warm-up, hidden size 256, and 8 attention heads.  
LLM rewrites are generated offline using DeepSeek-V3.2 with three style prompts (neutral, formal, sensational); detailed prompts are listed in Appendix~\ref{sec:prompt}.
Loss weights and hyperparameters are tuned on the validation set (Table~\ref{tab:hyperparams} in Appendix~\ref{sec:impl-details}). 
Once video/audio/text features are pre-extracted, the detector head reaches roughly $\sim$680 feature-level samples per second with batch size 128 on a single RTX 5090 GPU (about 1.5\,ms/sample and $\sim$1.67\,GB peak GPU memory on cached features); end-to-end latency is dominated by decoding and feature extraction, which can be amortized via caching and asynchronous preprocessing.
We have more efficiency and deployment discussion in Appendix~\ref{sec:efficiency} as well.

\subsection{Baselines}
\label{sec:baselines}

We compare \model{} against three categories of baselines. (i) \textit{Zero-shot VLMs}: GPT-4o-mini, GPT-4.1-mini, Qwen2.5-VL~\citep{bai2025qwen3}, InternVL2.5~\citep{chen2024expanding}, and InternVL2.5-MPO~\citep{wang2024enhancing}, evaluated by directly prompting them on the detection task. (ii) \textit{Supervised multimodal methods}: HCFC-Hou~\citep{bu2024fakingrecipe}, HCFC-Medina~\citep{serrano2020nlp} (hand-crafted feature classifiers), TikTec~\citep{shang2021multimodal}, FANVM~\citep{choi2021using}, SVFEND~\citep{qi2023fakesv}, FakingRecipe~\citep{bu2024fakingrecipe}, and KDSGAT-FNVD~\citep{huang2025knowledge}. (iii) \textit{VLM-based detector}: FakeSV-VLM~\citep{wang2025fakesv}, a fine-tuned VLM specifically designed for fake news video detection.

\subsection{Main Results}

\begin{table*}[t]
  \centering
  \footnotesize
  \setlength{\tabcolsep}{5pt}
  \begin{tabular}{lcccccccccccc}
    \toprule
    \multirow{3}{*}{\raisebox{-0.4ex}{\textbf{Method}}} &
    \multicolumn{6}{c}{\textbf{FakeSV}} &
    \multicolumn{6}{c}{\textbf{FakeTT}} \\
    \cmidrule(lr){2-7}\cmidrule(lr){8-13}
    & \multirow{2}{*}{\raisebox{-0.2ex}{\textbf{Acc}}} &
      \multirow{2}{*}{\raisebox{-0.2ex}{\textbf{macF1}}} &
      \multicolumn{2}{c}{\textbf{Real}} &
      \multicolumn{2}{c}{\textbf{Fake}} &
      \multirow{2}{*}{\raisebox{-0.2ex}{\textbf{Acc}}} &
      \multirow{2}{*}{\raisebox{-0.2ex}{\textbf{macF1}}} &
      \multicolumn{2}{c}{\textbf{Real}} &
      \multicolumn{2}{c}{\textbf{Fake}} \\
    \cmidrule(lr){4-5}\cmidrule(lr){6-7}\cmidrule(lr){10-11}\cmidrule(lr){12-13}
    & & & Pre & Rec & Pre & Rec & & & Pre & Rec & Pre & Rec \\
    \midrule
    \multicolumn{13}{l}{\textit{Zero-shot VLMs}} \\
    GPT-4o-mini &
      68.08 & 68.05 & 65.35 & 73.97 & 74.41 & 64.93 &
      61.54 & 61.20 & 81.62 & 52.00 & 47.28 & 79.80 \\
    GPT-4.1-mini &
      70.30 & 70.25 & 68.12 & 73.53 & 73.10 & 68.21 &
      49.16 & 48.54 & 77.42 & 48.00 & 37.50 & 71.40 \\
    Qwen2.5-VL &
      64.21 & 60.79 & 62.50 & 60.54 & 66.60 & 62.50 &
      45.82 & 45.31 & 72.10 & 45.00 & 41.28 & 65.84 \\
    InternVL2.5 &
      64.39 & 57.89 & 63.80 & 57.50 & 73.24 & 63.50 &
      46.82 & 45.29 & 77.60 & 49.00 & 52.24 & 69.46 \\
    InternVL2.5-MPO &
      65.13 & 61.07 & 63.75 & 59.16 & 69.17 & 65.08 &
      43.14 & 40.84 & 74.00 & 47.00 & 49.80 & 65.46 \\
    \midrule
    \multicolumn{13}{l}{\textit{Supervised baselines}} \\
    HCFC-Hou &
      74.91 & 73.61 & 77.72 & 60.08 & 73.46 & 86.51 &
      73.24 & 72.00 & 87.04 & 70.50 & 56.93 & 78.79 \\
    HCFC-Medina &
      76.38 & 75.83 & 74.77 & 69.75 & 77.50 & 81.58 &
      62.54 & 62.23 & 84.92 & 53.50 & 46.24 & 80.81 \\
    TikTec &
      73.43 & 73.26 & 68.08 & 74.37 & 78.37 & 72.70 &
      66.22 & 65.08 & 82.35 & 63.00 & 49.32 & 72.73 \\
    FANVM &
      79.52 & 78.81 & 80.98 & 69.75 & 78.64 & 87.17 &
      71.57 & 70.21 & 85.28 & 69.50 & 55.15 & 75.76 \\
    SVFEND &
      80.88 & 80.17 & 74.53 & 83.61 & 85.82 & 77.63 &
      77.14 & 75.63 & 87.91 & 76.33 & 62.33 & 78.79 \\
    FakingRecipe &
      82.87 & 83.52 & \best{91.35} & 71.00 & 80.67 & \best{94.74} &
      79.26 & 77.87 & 89.66 & 78.00 & 64.80 & 81.82 \\
    KDSGAT-FNVD &
      84.73 & 84.68 & 82.23 & 83.61 & 87.00 & 85.86 &
      81.94 & 78.94 & \second{92.68} & 76.00 & 64.44 & 87.88 \\
    \midrule
    \multicolumn{13}{l}{\textit{Ours}} \\
    FakeSV-VLM &
      \second{89.60} & \second{89.40} & 89.20 & \second{90.10} &
      \second{90.50} & 88.00 &
      \second{88.90} & \second{87.60} & 88.20 & \second{86.50} &
      \second{86.60} & 89.10 \\
    \rowcolor{magicrow}
    \textbf{\model{} (ours)} &
      86.71 & 86.84 & 86.52 & 83.61 & 87.50 & \second{89.80} &
      84.95 & 82.28 & \best{94.12} & 80.00 & 68.99 & \second{89.90} \\
    \rowcolor{magicrow}
    \textbf{\model{}+VLM (ours)} &
      \best{90.93} & \best{90.41} & \second{90.85} & \best{90.95} &
      \best{91.10} & 88.90 &
      \best{89.52} & \best{88.14} & 89.70 & \best{88.50} &
      \best{87.20} & \best{90.10} \\
    \midrule
    \multicolumn{13}{l}{\textit{End-to-End Efficiency}} \\
    FakeSV-VLM &
      \multicolumn{6}{c}{18.67 samp/s @24.68 GB} &
      \multicolumn{6}{c}{30.25 samp/s @22.76 GB} \\
    \rowcolor{magicrow}
    \textbf{\model{} (ours)} &
      \multicolumn{6}{c}{\textbf{680.43 samp/s @1.67 GB}} &
      \multicolumn{6}{c}{\textbf{693.76 samp/s @1.67 GB}} \\
    \rowcolor{magicrow}
   \textbf{\model{}+VLM (ours)} &
      \multicolumn{6}{c}{\underline{509.71 samp/s @24.68 GB}} &
      \multicolumn{6}{c}{\underline{530.73 samp/s @22.76 GB}} \\
    \bottomrule
  \end{tabular}
  \caption{Main results on FakeSV and FakeTT. Best values in \best{bold}, second best \second{underlined}. \model{}+VLM routes 25\% of samples to VLM (Figure~\ref{fig:uncertainty-routing}) while achieving comparable or better accuracy, offering the \textbf{best cost--performance tradeoff}: VLM-level or higher accuracy at up to 18--27$\times$ higher throughput than VLM-only.}
  \label{tab:main_results}
\end{table*}

\begin{figure*}[t]
  \centering
  \begin{subfigure}{0.48\textwidth}
    \centering
    \includegraphics[width=\linewidth]{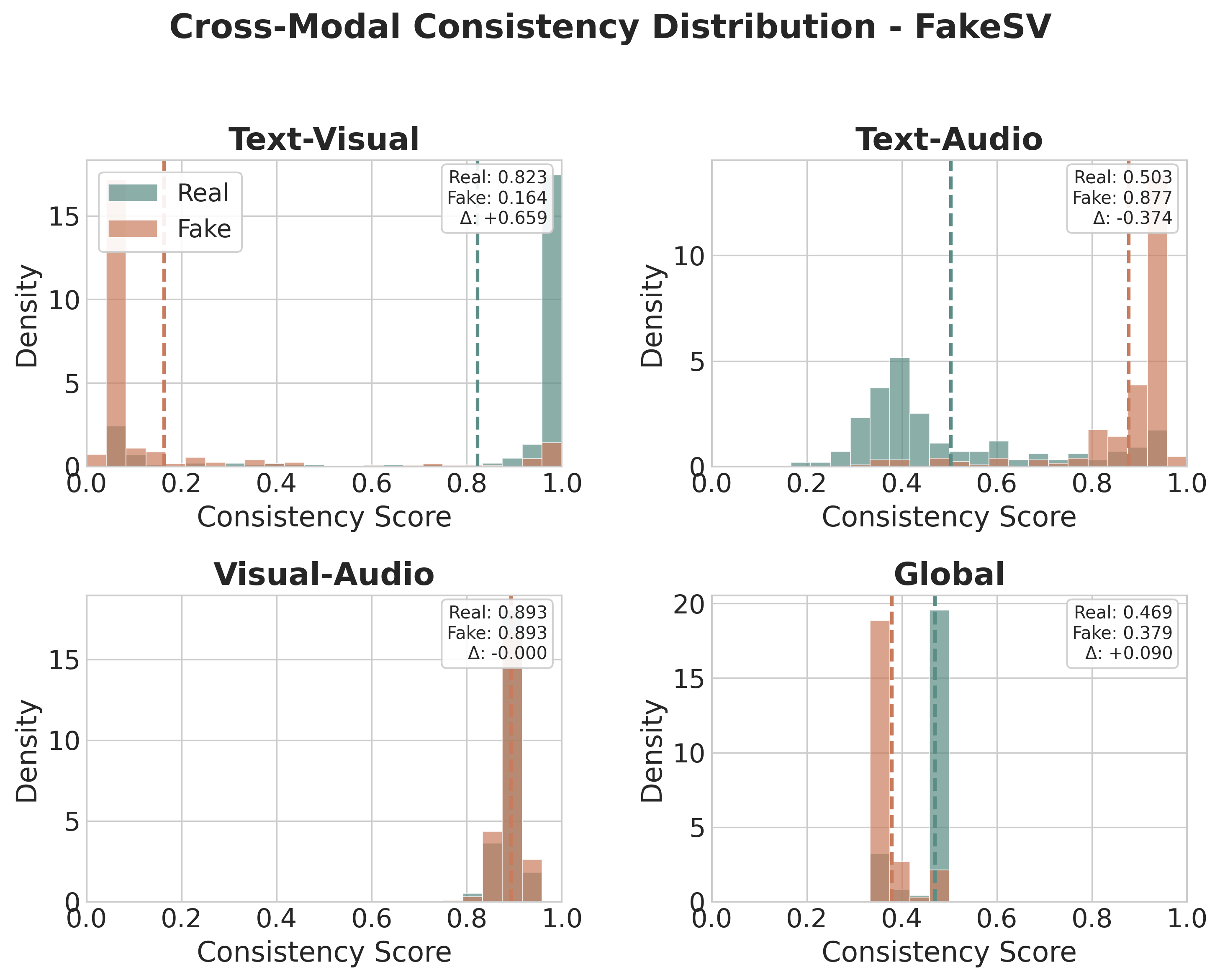}
  \end{subfigure}
  \hfill
  \begin{subfigure}{0.48\textwidth}
    \centering
    \includegraphics[width=\linewidth]{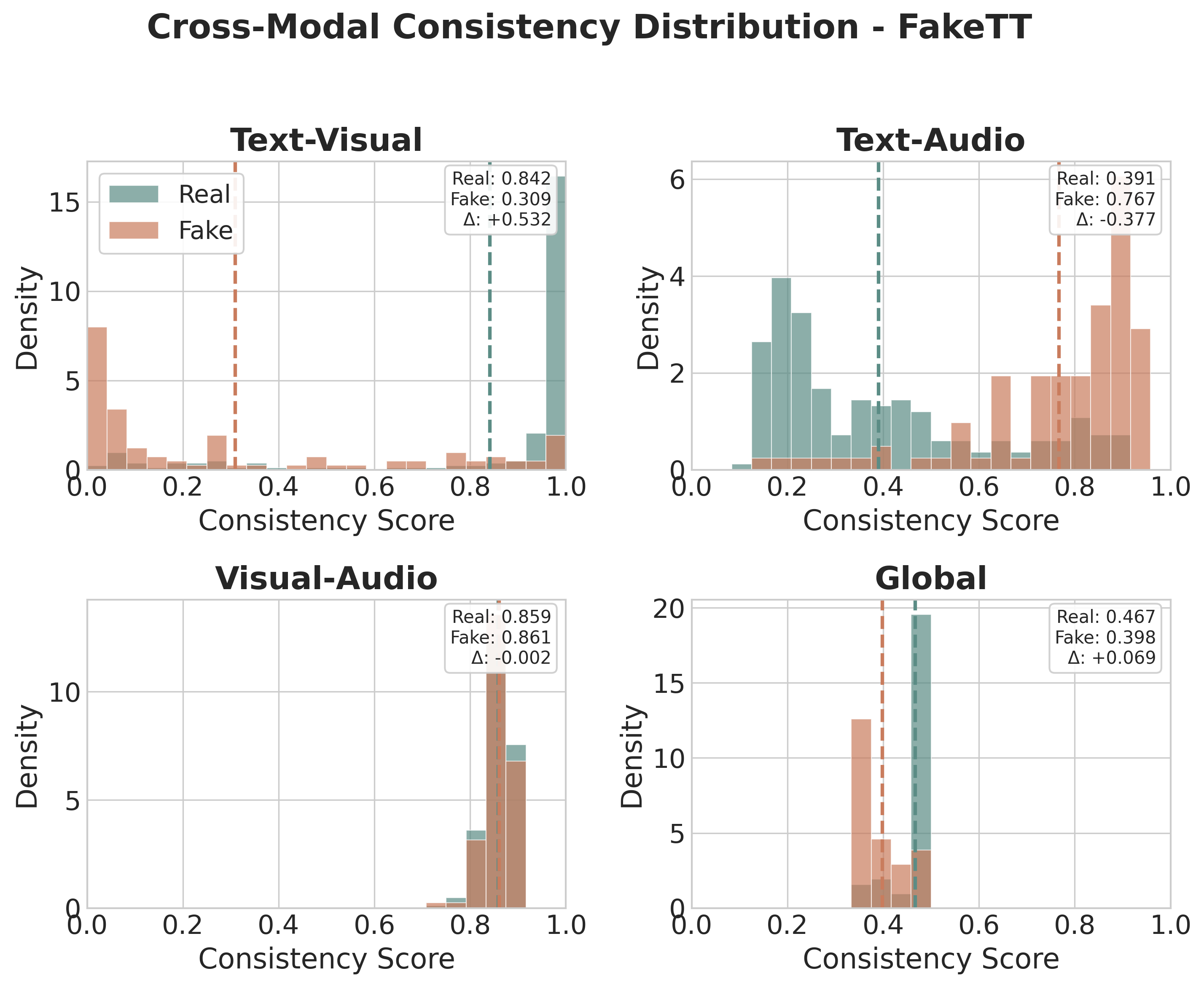}
  \end{subfigure}

  \caption{Cross-modal consistency distributions of text--visual, text--audio, visual--audio, and global consistency scores for real vs.\ fake videos. Real news shows high text--visual but moderate text--audio consistency; fake news flips this pattern, while visual--audio consistency remains high. Left: FakeSV; right: FakeTT.}
  \label{fig:consistency-distribution}
\end{figure*}


Table~\ref{tab:main_results} reports results on FakeTT and FakeSV.  
We compare \model{} with the baselines detailed in Section~\ref{sec:baselines}.
For each dataset, we report accuracy, macro-F1, and per-class precision/recall. On FakeSV, \model{} reaches 86.71\% accuracy and 86.84 macro-F1, outperforming KDSGAT-FNVD by 1.98 and 2.16 points, respectively, with balanced precision/recall across real and fake classes. On FakeTT, \model{} achieves 84.95\% accuracy and 82.28 macro-F1, improving over KDSGAT-FNVD by 3.01 and 3.34 points. Zero-shot VLMs lag far behind: on FakeSV the best VLM trails \model{} by about 16 accuracy points, and on FakeTT the gap exceeds 20 points, underscoring the value of supervision and explicit consistency modeling. These gains also hold across macro-F1 and per-class precision/recall, particularly on the real class of FakeTT where \model{} attains the highest precision.

When combined with a heavyweight VLM (FakeSV-VLM) in a two-stage pipeline, \model{}+VLM makes further improvement and even surpasses the specialized VLM-only detector FakeSV-VLM on both FakeSV (90.93\% vs. 89.60\%) and FakeTT (89.52\% vs. 88.90\%), while still clearly outperforming all non-VLM baselines (Table~\ref{tab:main_results}). A detailed analysis of the routing policy and its superior cost--performance benefits is provided in Section~\ref{sec:twostage}.

\subsection{Ablation Study}

\begin{table}[t]
  \centering
  \small
  \setlength{\tabcolsep}{4pt}
  \renewcommand{\arraystretch}{0.95}
  \begin{tabular*}{\columnwidth}{@{\extracolsep{\fill}}c l c c c@{}}
    \toprule
    & \textbf{Variant} & \textbf{Acc} & \textbf{F1} & \textbf{AUC} \\
    \midrule
    \multicolumn{5}{c}{\textbf{FakeSV}} \\
    \midrule
    & \model{} (full) & \best{86.71} & \best{86.84} & 92.23 \\
    \cmidrule{2-5}
    \multirow{2}{*}{Consistency}
      & w/o CMCG & 84.90 & 84.85 & 91.77 \\
      & w/o CFE  & 85.69 & 85.61 & \second{92.30} \\
      & w/o TCMI & 85.75 & 85.89 & 91.94 \\
    \cmidrule{2-5}
    \multirow{2}{*}{Fusion}
      & w/o AARF & 84.49 & 84.71 & 91.43 \\
      & w/o HMT  & 85.86 & 85.79 & \best{92.48} \\
    \cmidrule{2-5}
    \multirow{2}{*}{Training}
      & w/o contr. & 85.75 & 85.89 & 92.15 \\
      & w/o adv.   & \second{86.56} & \second{86.53} & 92.23 \\
    \midrule
    \multicolumn{5}{c}{\textbf{FakeTT}} \\
    \midrule
    & \model{} (full) & \best{84.95} & \best{82.28} & \best{89.08} \\
    \cmidrule{2-5}
    \multirow{2}{*}{Consistency}
      & w/o CMCG & 81.94 & 78.68 & 87.44 \\
      & w/o CFE  & 78.83 & 80.13 & 87.88 \\
      & w/o TCMI & 78.64 & 77.22 & 86.92 \\
    \cmidrule{2-5}
    \multirow{2}{*}{Fusion}
      & w/o AARF & 80.91 & 78.90 & 88.30 \\
      & w/o HMT  & 82.91 & \second{81.48} & 88.47 \\
    \cmidrule{2-5}
    \multirow{2}{*}{Training}
      & w/o contr. & \second{82.93} & 80.47 & \second{88.50} \\
      & w/o adv.   & 78.37 & 77.78 & 88.37 \\
    \bottomrule
  \end{tabular*}
  \caption{Ablation study. Consistency modules (CMCG, CFE, TCMI) and AARF are critical.}
  \label{tab:ablation}
\end{table}

\begin{figure*}[t]
  \centering
  \begin{subfigure}{0.48\textwidth}
    \centering
    \includegraphics[width=\linewidth]{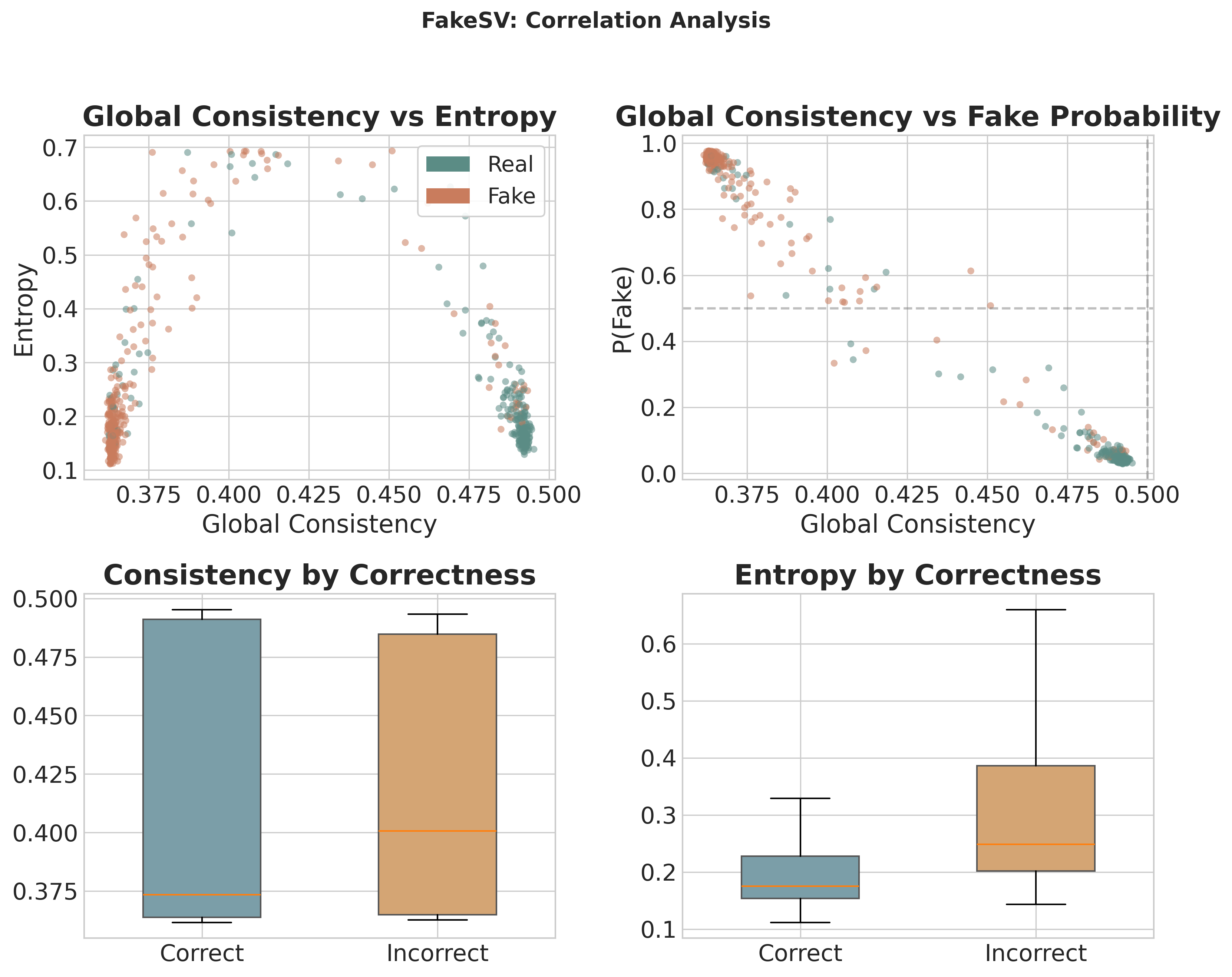}
  \end{subfigure}
  \hfill
  \begin{subfigure}{0.48\textwidth}
    \centering
    \includegraphics[width=\linewidth]{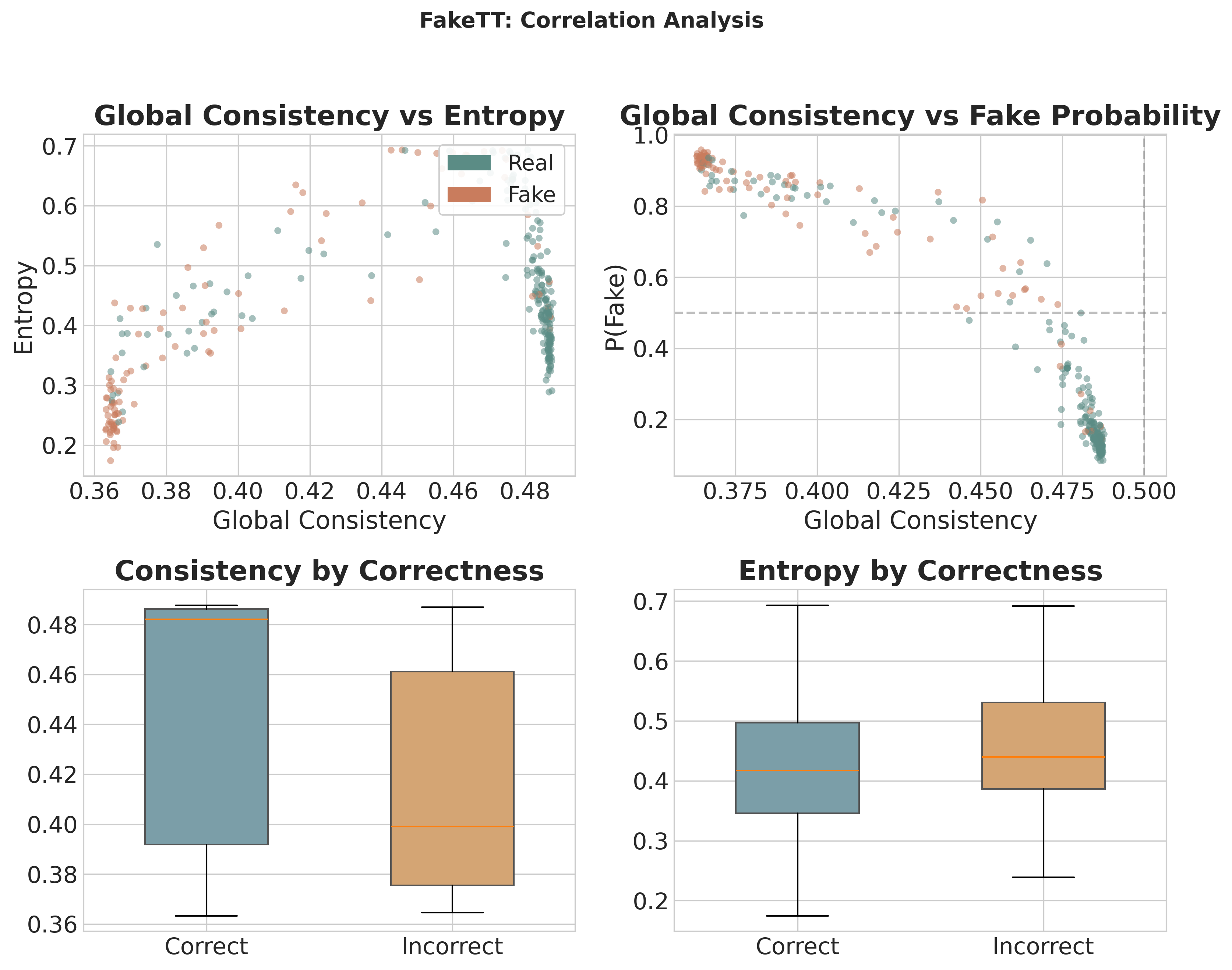}
  \end{subfigure}
  \caption{Consistency--prediction analysis. Global consistency is strongly (negatively) correlated with fake probability; errors concentrate in the middle-consistency band, while $c_{\mathrm{global}}$ vs.\ entropy is non-linear with uncertainty peaking at mid-range consistency. Left: FakeSV; right: FakeTT.}
  \label{fig:cfe-qualitative}
\end{figure*}

To assess the contribution of each component, we conduct ablations on both datasets (Table~\ref{tab:ablation}).  
Removing core consistency modules (CMCG, CFE, TCMI) consistently degrades performance, confirming that explicit consistency modeling is effective.  
AARF style-robust fusion impacts performance particularly on FakeTT, supporting the need for handling style shifts.
CFE and TCMI also contribute to interpretability, driving the field visualizations and temporal inconsistency detection.

\section{Analysis and Findings}
\label{sec:findings}

Beyond the main results, we investigate the consistency signals to uncover deeper insights about how real and fake videos differ.

\subsection{Asymmetric Cross-Modal Consistency}
We first examine cross-modal consistency in aggregate for real vs.\ fake videos. Using CMCG, we compute four scalar scores per video: text--visual, text--audio, visual--audio, and global consistency. Figure~\ref{fig:consistency-distribution} shows their distributions for FakeSV (left) and FakeTT (right). In both datasets, real videos exhibit high text--visual and moderate text--audio consistency, whereas fake videos show the reverse: text--audio consistency is very high, but text--visual consistency collapses. Visual--audio consistency remains high for both. Global consistency is, on average, higher for real videos, but the asymmetry is the distinguishing factor.

\paragraph{Underlying Causes of Asymmetric Cross-Modal Consistency}
This asymmetry is consistent with the production mechanics of ``cheapfakes''~\citep{paris2019deepfakes}.
Qualitatively, we observe that fake news producers often start with high-quality, visually coherent stock footage or stolen clips, which naturally preserves strong alignment between the visual stream and background audio.
These materials are then repurposed with fabricated subtitles and voiceovers to convey a false narrative.
The generated text is designed to be emotionally engaging and closely aligned with the affect of the audio, while gradually drifting away from the actual visual evidence, creating the semantic gap that \model{} detects.
In contrast, real news videos, particularly those from professional outlets in FakeTT, emphasize factual grounding, with voiceovers that closely describe the visual events, resulting in stronger text--visual consistency.
More statistics are shown in Appendix~\ref{subsec:agg-stats}.

\subsection{Global Consistency for Interpretability}

We examine how the global consistency score $c_{\mathrm{global}}$ relates to predictions.  
Figure~\ref{fig:cfe-qualitative} plots $c_{\mathrm{global}}$ against the predicted fake probability on FakeSV (left) and FakeTT (right).  
$c_{\mathrm{global}}$ is strongly and roughly \emph{negatively} correlated with the model's fake probability; prediction errors cluster in the middle band of $c_{\mathrm{global}}$, indicating that the score provides an interpretable, low-dimensional difficulty axis.  

\paragraph{Interpretability of Global Consistency}
Because $c_{\mathrm{global}}$ is an explicit input feature, this correlation is expected. However, its value lies in its \emph{operational utility}. 
Unlike opaque end-to-end probabilities, $c_{\mathrm{global}}$ offers human moderators a tangible explanation for the model's decision: a video is flagged not just because it ``looks fake,'' but specifically because its modalities fail to cohere. 
We show that uncertainty peaks precisely in the middle-consistency region in Appendix~\ref{subsec:cglobal-corr}, suggesting that $c_{\mathrm{global}}$ can serve as a filter to identify ambiguous cases that require human or VLM intervention.

\subsection{Uncertainty-Aware Two-Stage VLM Routing}

\label{sec:twostage}

We now examine how consistency and uncertainty can be used to orchestrate a two-stage pipeline with a heavyweight VLM detector.  
Figure~\ref{fig:uncertainty-routing} summarizes uncertainty distributions, calibration curves, and routing behavior on both datasets. We implement FakeSV-VLM~\citep{wang2025fakesv} for the stage-2 detector. The top row shows that entropy is typically lower for correct predictions than for incorrect ones, and that confidence is reasonably calibrated.  
The bottom row then visualizes routing policies that identify difficult samples for VLM escalation.  
We treat the routing threshold as a hyperparameter and tune it on the validation set to route approximately 25\% of samples on both datasets.

Routed samples are indeed harder (lower detector-only accuracy), and the overall accuracy of the two-stage system improves.  
As shown in Table~\ref{tab:main_results}, on FakeSV, routing 25\% of samples to a VLM (\model{}+VLM) increases overall accuracy from 86.71\% to 90.93\%, surpassing FakeSV-VLM's 89.60\% by 1.33 percentage points.  
On FakeTT, a 25\% routing ratio boosts accuracy from 84.95\% to 89.52\%, surpassing the VLM-only baseline of 88.90\% by 0.62pp.  
Crucially, this hybrid approach achieves 18--27$\times$ higher throughput and massive VRAM savings compared to a pure VLM pipeline. \model{} effectively filters out the bulk of easy cases ($\sim$75\%), allowing expensive VLMs to focus only on the hardest samples, and thus achieving the best cost--performance tradeoff.

\subsection{Consistency Stability under Style Perturbations}

\label{sec:style-finding}

Finally, we explore the effect of multi-style LLM rewrites.  
Adversarial-Aware Rewrite Fusion fuses the original caption with neutral, formal, and sensational rewrites and is trained with style-aware contrastive losses.  
Table~\ref{tab:ablation} reports an ablation in which we remove AARF (i.e., use only the original text) while keeping other components fixed.  
Performance drops on both datasets, indicating that exposure to style variation improves robustness. 
Conceptually, this connects to work on style-aware neural fake news detection like Style-News~\citep{wang2024style}, where stylized generation is used as an adversary to stress-test detectors.  
Our setup differs in two ways: we treat rewrites as semantic-preserving perturbations generated by a general-purpose LLM (rather than a task-specific generator), and we explicitly measure how cross-modal consistency behaves under style changes (statistical details in Appendix~\ref{subsec:style-stats}).
Our hypothesis is that real news remains consistent under style rewrites, whereas fake news shows larger variance, especially in text--visual consistency.  
This is supported qualitatively by Figure~\ref{fig:style-heatmap}, where fake samples exhibit larger shifts under stylistic perturbations.  
Appendix~\ref{subsec:style-stats} reports detailed variance statistics.

\section{Conclusion}


We presented \model{}, a consistency-centric detector for fake news short videos that explicitly exposes cross-tri-modal consistency at multiple granularities: pairwise and global scores, token-level consistency fields, a temporal audio--visual inconsistency score, and uncertainty estimates. 
On FakeSV and FakeTT, \model{} achieves state-of-the-art supervised performance on frozen features and offers a superior cost--performance tradeoff: when combined with VLM routing, it surpasses VLM-only accuracy at 18--27$\times$ higher throughput.
Future work should integrate external knowledge search for fact-level verification.

\section*{Limitations}
First, our method relies on pre-extracted multimodal features and offline LLM-generated text rewrites, which introduce additional pre-processing cost and potential sensitivity to the choice of LLMs and prompt designs; however, this is a deliberate design choice that decouples heavy backbone computation from the detector head, enabling high throughput, low memory usage, and flexible integration with different pretrained encoders, and our ablation results further show that the core consistency modeling remains effective even without LLM rewrites.
Second, the current TCMI module focuses on coarse-grained temporal audio–visual inconsistency and does not explicitly model fine-grained temporal logic or causal structures; nevertheless, it is intentionally designed as a lightweight complement to static consistency signals, targeting prominent synchronization mismatches commonly observed in repurposed or cheaply manipulated videos.
Third, \model{} operates as a feature-level detector and does not directly process raw video inputs; while this limits strictly on-device, end-to-end deployment, it allows the framework to remain backbone-agnostic and deployment-friendly for large-scale platform settings where features are often cached or shared across tasks.
Moreover, consistency-based detectors may still struggle with carefully crafted misinformation in which all modalities are internally consistent but factually incorrect; addressing such cases typically requires external knowledge and verification resources, and we view our consistency-centric approach as a complementary detection axis rather than a standalone solution to factual verification.
Finally, although the proposed consistency fields and uncertainty estimates improve model transparency and enable fine-grained analysis, more human-centered evaluations are needed to fully understand how practitioners interpret and act upon these signals, which we leave as an important direction for future work.

\section*{Ethics Statement}

This work aims to mitigate the harmful impact of multimedia misinformation in short-form news videos.  
Nonetheless, there are dual-use risks: understanding detection mechanisms can inform adversaries seeking to evade them.  
We therefore focus on general architectural principles rather than attack-specific heuristics, and we encourage responsible deployment practices that combine automated detection with human oversight.

All datasets used are publicly available; we follow their original licenses and respect existing privacy safeguards.  
In real-world deployments, particular care should be taken to monitor false positives, ensure fairness across topics and languages, and provide appropriate channels for appeals when automated decisions affect users.

\bibliography{custom}

\begin{thebibliography}{28}
\providecommand{\natexlab}[1]{#1}

\bibitem[{Amiriparian et~al.(2024)Amiriparian, Packa{\'n}, Gerczuk, and Schuller}]{amiriparian2024exhubert}
Shahin Amiriparian, Filip Packa{\'n}, Maurice Gerczuk, and Bj{\"o}rn~W Schuller. 2024.
\newblock Exhubert: Enhancing hubert through block extension and fine-tuning on 37 emotion datasets.
\newblock \emph{arXiv preprint arXiv:2406.10275}.

\bibitem[{Bai et~al.(2025)Bai, Cai, Chen, Chen, Chen, Cheng, Deng, Ding, Gao, Ge et~al.}]{bai2025qwen3}
Shuai Bai, Yuxuan Cai, Ruizhe Chen, Keqin Chen, Xionghui Chen, Zesen Cheng, Lianghao Deng, Wei Ding, Chang Gao, Chunjiang Ge, and 1 others. 2025.
\newblock Qwen3-vl technical report.
\newblock \emph{arXiv preprint arXiv:2511.21631}.

\bibitem[{Bu et~al.(2023)Bu, Sheng, Cao, Qi, Wang, and Li}]{bu2023combating}
Yuyan Bu, Qiang Sheng, Juan Cao, Peng Qi, Danding Wang, and Jintao Li. 2023.
\newblock Combating online misinformation videos: Characterization, detection, and future directions.
\newblock In \emph{Proceedings of the 31st ACM International Conference on Multimedia}, pages 8770--8780.

\bibitem[{Bu et~al.(2024)Bu, Sheng, Cao, Qi, Wang, and Li}]{bu2024fakingrecipe}
Yuyan Bu, Qiang Sheng, Juan Cao, Peng Qi, Danding Wang, and Jintao Li. 2024.
\newblock Fakingrecipe: Detecting fake news on short video platforms from the perspective of creative process.
\newblock In \emph{Proceedings of the 32nd ACM International Conference on Multimedia}, pages 1351--1360.

\bibitem[{Chen et~al.(2024)Chen, Wang, Cao, Liu, Gao, Cui, Zhu, Ye, Tian, Liu et~al.}]{chen2024expanding}
Zhe Chen, Weiyun Wang, Yue Cao, Yangzhou Liu, Zhangwei Gao, Erfei Cui, Jinguo Zhu, Shenglong Ye, Hao Tian, Zhaoyang Liu, and 1 others. 2024.
\newblock Expanding performance boundaries of open-source multimodal models with model, data, and test-time scaling.
\newblock \emph{arXiv preprint arXiv:2412.05271}.

\bibitem[{Choi and Ko(2021)}]{choi2021using}
Hyewon Choi and Youngjoong Ko. 2021.
\newblock Using topic modeling and adversarial neural networks for fake news video detection.
\newblock In \emph{Proceedings of the 30th ACM international conference on information \& knowledge management}, pages 2950--2954.

\bibitem[{Devlin et~al.(2019)Devlin, Chang, Lee, and Toutanova}]{devlin2019bert}
Jacob Devlin, Ming-Wei Chang, Kenton Lee, and Kristina Toutanova. 2019.
\newblock Bert: Pre-training of deep bidirectional transformers for language understanding.
\newblock In \emph{Proceedings of the 2019 conference of the North American chapter of the association for computational linguistics: human language technologies, volume 1 (long and short papers)}, pages 4171--4186.

\bibitem[{Gao et~al.(2024)Gao, Wang, Zhang, Zeng, and Ma}]{gao2024temporal}
Yuan Gao, Xuelong Wang, Yu~Zhang, Ping Zeng, and Yingjie Ma. 2024.
\newblock Temporal feature prediction in audio--visual deepfake detection.
\newblock \emph{Electronics}, 13(17):3433.

\bibitem[{Huang et~al.(2025)Huang, Ma, Tang, and Rong}]{huang2025knowledge}
Xuejian Huang, Tinghuai Ma, Hao Tang, and Huan Rong. 2025.
\newblock Knowledge-enhanced dynamic scene graph attention network for fake news video detection.
\newblock \emph{IEEE Transactions on Multimedia}.

\bibitem[{Li et~al.(2025)Li, Qiao, Yin, Wu, Gao, Wang, and Li}]{li2025survey}
Xianghua Li, Jiao Qiao, Shu Yin, Lianwei Wu, Chao Gao, Zhen Wang, and Xuelong Li. 2025.
\newblock A survey of multimodal fake news detection: a cross-modal interaction perspective.
\newblock \emph{IEEE Transactions on Emerging Topics in Computational Intelligence}.

\bibitem[{Liu et~al.(2021)Liu, Lin, Cao, Hu, Wei, Zhang, Lin, and Guo}]{liu2021swin}
Ze~Liu, Yutong Lin, Yue Cao, Han Hu, Yixuan Wei, Zheng Zhang, Stephen Lin, and Baining Guo. 2021.
\newblock Swin transformer: Hierarchical vision transformer using shifted windows.
\newblock In \emph{Proceedings of the IEEE/CVF international conference on computer vision}, pages 10012--10022.

\bibitem[{M{\"u}ller-Budack et~al.(2020)M{\"u}ller-Budack, Theiner, Diering, Idahl, and Ewerth}]{muller2020multimodal}
Eric M{\"u}ller-Budack, Jonas Theiner, Sebastian Diering, Maximilian Idahl, and Ralph Ewerth. 2020.
\newblock Multimodal analytics for real-world news using measures of cross-modal entity consistency.
\newblock In \emph{Proceedings of the 2020 international conference on multimedia retrieval}, pages 16--25.

\bibitem[{Paris and Donovan(2019)}]{paris2019deepfakes}
Britt Paris and Joan Donovan. 2019.
\newblock Deepfakes and cheap fakes.
\newblock \emph{United States of America: Data \& Society}, 1.

\bibitem[{Qi et~al.(2023)Qi, Bu, Cao, Ji, Shui, Xiao, Wang, and Chua}]{qi2023fakesv}
Peng Qi, Yuyan Bu, Juan Cao, Wei Ji, Ruihao Shui, Junbin Xiao, Danding Wang, and Tat-Seng Chua. 2023.
\newblock Fakesv: A multimodal benchmark with rich social context for fake news detection on short video platforms.
\newblock In \emph{Proceedings of the AAAI Conference on Artificial Intelligence}, volume~37, pages 14444--14452.

\bibitem[{Serrano et~al.(2020)Serrano, Papakyriakopoulos, and Hegelich}]{serrano2020nlp}
Juan Carlos~Medina Serrano, Orestis Papakyriakopoulos, and Simon Hegelich. 2020.
\newblock Nlp-based feature extraction for the detection of covid-19 misinformation videos on youtube.
\newblock In \emph{Proceedings of the 1st Workshop on NLP for COVID-19 at ACL 2020}.

\bibitem[{Shang et~al.(2021)Shang, Kou, Zhang, and Wang}]{shang2021multimodal}
Lanyu Shang, Ziyi Kou, Yang Zhang, and Dong Wang. 2021.
\newblock A multimodal misinformation detector for covid-19 short videos on tiktok.
\newblock In \emph{2021 IEEE international conference on big data (big data)}, pages 899--908. IEEE.

\bibitem[{Shu et~al.(2019)Shu, Wang, and Liu}]{shu2019beyond}
Kai Shu, Suhang Wang, and Huan Liu. 2019.
\newblock Beyond news contents: The role of social context for fake news detection.
\newblock In \emph{Proceedings of the twelfth ACM international conference on web search and data mining}, pages 312--320.

\bibitem[{Tian et~al.(2025)Tian, Ho, and Chen}]{tian2025symbolic}
Chong Tian, Qirong Ho, and Xiuying Chen. 2025.
\newblock A symbolic adversarial learning framework for evolving fake news generation and detection.
\newblock In \emph{Proceedings of the 2025 Conference on Empirical Methods in Natural Language Processing}, pages 12307--12321.

\bibitem[{Vosoughi et~al.(2018)Vosoughi, Roy, and Aral}]{vosoughi2018spread}
Soroush Vosoughi, Deb Roy, and Sinan Aral. 2018.
\newblock The spread of true and false news online.
\newblock \emph{science}, 359(6380):1146--1151.

\bibitem[{Wang et~al.(2025{\natexlab{a}})Wang, Yang, and Zhong}]{wang2025global}
Haoran Wang, Yan Yang, and Yingli Zhong. 2025{\natexlab{a}}.
\newblock Global and local feature enhancement for short video fake news detection.
\newblock In \emph{International Conference on Intelligent Computing}, pages 435--446. Springer.

\bibitem[{Wang et~al.(2025{\natexlab{b}})Wang, Liu, Zhang, and Wang}]{wang2025consistency}
Junxi Wang, Jize Liu, Na~Zhang, and Yaxiong Wang. 2025{\natexlab{b}}.
\newblock Consistency-aware fake videos detection on short video platforms.
\newblock In \emph{International Conference on Intelligent Computing}, pages 200--210. Springer.

\bibitem[{Wang et~al.(2025{\natexlab{c}})Wang, Wang, Cheng, and Zhong}]{wang2025fakesv}
Junxi Wang, Yaxiong Wang, Lechao Cheng, and Zhun Zhong. 2025{\natexlab{c}}.
\newblock Fakesv-vlm: Taming vlm for detecting fake short-video news via progressive mixture-ofexperts adapter.
\newblock \emph{arXiv preprint arXiv:2508.19639}.

\bibitem[{Wang et~al.(2024{\natexlab{a}})Wang, Chang, and Peng}]{wang2024style}
Wei-Yao Wang, Yu-Chieh Chang, and Wen-Chih Peng. 2024{\natexlab{a}}.
\newblock Style-news: Incorporating stylized news generation and adversarial verification for neural fake news detection.
\newblock \emph{arXiv preprint arXiv:2401.15509}.

\bibitem[{Wang et~al.(2024{\natexlab{b}})Wang, Chen, Wang, Cao, Liu, Gao, Zhu, Zhu, Lu, Qiao et~al.}]{wang2024enhancing}
Weiyun Wang, Zhe Chen, Wenhai Wang, Yue Cao, Yangzhou Liu, Zhangwei Gao, Jinguo Zhu, Xizhou Zhu, Lewei Lu, Yu~Qiao, and 1 others. 2024{\natexlab{b}}.
\newblock Enhancing the reasoning ability of multimodal large language models via mixed preference optimization.
\newblock \emph{arXiv preprint arXiv:2411.10442}.

\bibitem[{Wang et~al.(2018)Wang, Ma, Jin, Yuan, Xun, Jha, Su, and Gao}]{wang2018eann}
Yaqing Wang, Fenglong Ma, Zhiwei Jin, Ye~Yuan, Guangxu Xun, Kishlay Jha, Lu~Su, and Jing Gao. 2018.
\newblock Eann: Event adversarial neural networks for multi-modal fake news detection.
\newblock In \emph{Proceedings of the 24th acm sigkdd international conference on knowledge discovery \& data mining}, pages 849--857.

\bibitem[{Yang et~al.(2025)Yang, Shi, Li, Fan, and Xu}]{yang2025fake}
Yukun Yang, Xiwei Shi, Haoxu Li, Buwei Fan, and Yijia Xu. 2025.
\newblock Fake news detection in short videos by integrating semantic credibility and multi-granularity contrastive learning.
\newblock \emph{Applied Sciences}, 15(23):12621.

\bibitem[{Zhou et~al.(2020)Zhou, Wu, and Zafarani}]{zhou2020safe}
X~Zhou, J~Wu, and R~Zafarani. 2020.
\newblock Safe: similarity-aware multi-modal fake news detection. arxiv.
\newblock \emph{arXiv preprint arXiv:2003.04981}.

\bibitem[{Zhou and Zafarani(2020)}]{zhou2020survey}
Xinyi Zhou and Reza Zafarani. 2020.
\newblock A survey of fake news: Fundamental theories, detection methods, and opportunities.
\newblock \emph{ACM Computing Surveys (CSUR)}, 53(5):1--40.

\end{thebibliography}

\clearpage
\appendix

\section{Cross-modal Inconsistency Taxonomy (Details)}
\label{sec:taxonomy-appendix}

Table~\ref{tab:taxonomy} provides a detailed version of the task-oriented inconsistency taxonomy mentioned in the introduction.  
For each category we list a brief description and typical patterns; \model{}'s corresponding signals are discussed in the main text.

\begin{table*}[t]
  \centering
  \small
  \begin{tabular}{p{3cm}p{13.2cm}}
    \toprule
    \textbf{Category} & \textbf{Description and typical patterns} \\
    \midrule
    Entity \& context inconsistency &
    Entities (persons, organizations, places) or contexts (e.g., country, city, venue) mentioned in text do not match those present in video or audio. A typical pattern is subtitles claiming ``thousands protest in Paris'' while the video shows a small rally in a different city. \\
    Event-level semantic inconsistency &
    Modalities disagree on the core event: what happened, at what scale, or with which participants. For example, the caption claims ``violent clashes erupt'' but the video shows a peaceful gathering, or the audio describes a different incident. \\
    Affective inconsistency &
    Sentiment or emotion conveyed by audio contradicts text or visual content, such as tragic subtitles overlaid on upbeat background music or calm narration over highly dramatic visuals. \\
    Temporal audio--visual inconsistency &
    Audio and visual streams are misaligned in time, reused from different footage, or re-ordered. This includes dubbed videos, re-edited footage, or mis-timed sound effects. \\
    Stylistic / pragmatic inconsistency &
    Rhetorical style, framing, or level of sensationalism in one modality is inconsistent with the others, e.g., a sensational subtitle for a relatively mundane video, or highly emotional text with neutral audio. \\
    \bottomrule
  \end{tabular}
  \caption{Task-oriented taxonomy of inconsistencies in short-form news videos. These categories serve as design targets for \model{}'s consistency-centric architecture.}
  \label{tab:taxonomy}
\end{table*}

\begin{table*}[t]
  \centering
  \small
  \begin{tabular}{p{1.7cm}p{4.1cm}p{3.0cm}p{4.4cm}}
    \toprule
    \textbf{Component} & \textbf{Function} & \textbf{Consistency level} & \textbf{Role in findings / interpretability} \\
    \midrule
    \multicolumn{4}{l}{\textit{Consistency Computation}} \\
    CMCG &
    Computes pairwise consistency scores $[c_{\mathrm{tv}},c_{\mathrm{ta}},c_{\mathrm{va}}]$ and global score $c_{\mathrm{global}}$ &
    Pairwise $\rightarrow$ global &
    Reveals asymmetric text--visual vs.\ text--audio patterns; defines the global consistency axis (Figures~\ref{fig:consistency-distribution}--\ref{fig:cfe-qualitative}). \\
    CFE &
    Derives token/frame-level consistency fields from cross-modal attention &
    Local (token / frame) &
    Localizes suspicious phrases or frames; supports heatmap visualization (Figure~\ref{fig:cfe-fields}). \\
    TCMI &
    Computes temporal audio--visual inconsistency score $c_{\mathrm{temp}}$ &
    Temporal &
    Detects coarse audio--visual synchronization mismatches common in repurposed videos. \\
    \midrule
    \multicolumn{4}{l}{\textit{Cross-Modal Fusion}} \\
    AARF &
    Fuses original text with neutral/formal/sensational LLM rewrites via gated weights &
    Text-internal &
    Builds style-robust text embeddings; used to study style perturbation effects (Finding~4). \\
    HMT &
    Three-stage hierarchical fusion: intra-modal refinement, consistency-weighted cross-attention, global aggregation &
    Structural &
    Integrates multimodal representations with consistency-modulated attention. \\
    \midrule
    \multicolumn{4}{l}{\textit{Training and Prediction}} \\
    CAJL &
    Contrastive--adversarial joint learning with InfoNCE and perturbation losses &
    Training &
    Aligns multimodal representations and improves robustness under perturbations. \\
    Classifier &
    Outputs fake probability, confidence, and uncertainty for two-stage routing &
    Prediction &
    Provides calibrated uncertainty for selective VLM escalation (Finding~3). \\
    \bottomrule
  \end{tabular}
  \caption{\model{} component overview. Components are grouped by function: consistency computation (CMCG, CFE, TCMI), cross-modal fusion (AARF, HMT), and training/prediction (CAJL, Classifier). Together they provide the signals used in our four findings and the two-stage VLM pipeline.}
  \label{tab:magic_pipeline}
\end{table*}

\section{Additional Figures}
\label{sec:addl-figures}
\begin{figure*}[t]
  \centering
  \begin{subfigure}{0.48\textwidth}
    \centering
    \includegraphics[width=\linewidth]{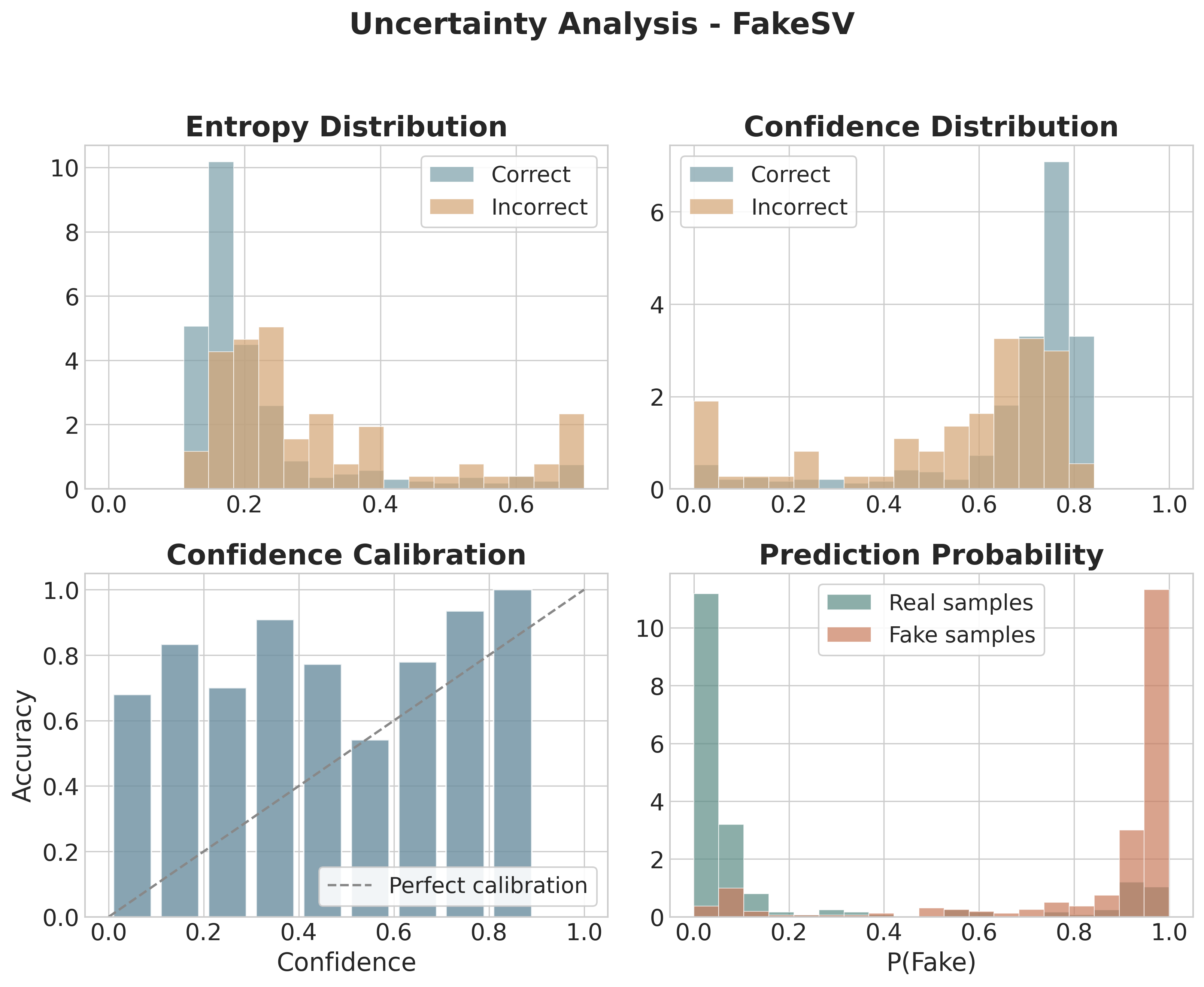}
  \end{subfigure}
  \hfill
  \begin{subfigure}{0.48\textwidth}
    \centering
    \includegraphics[width=\linewidth]{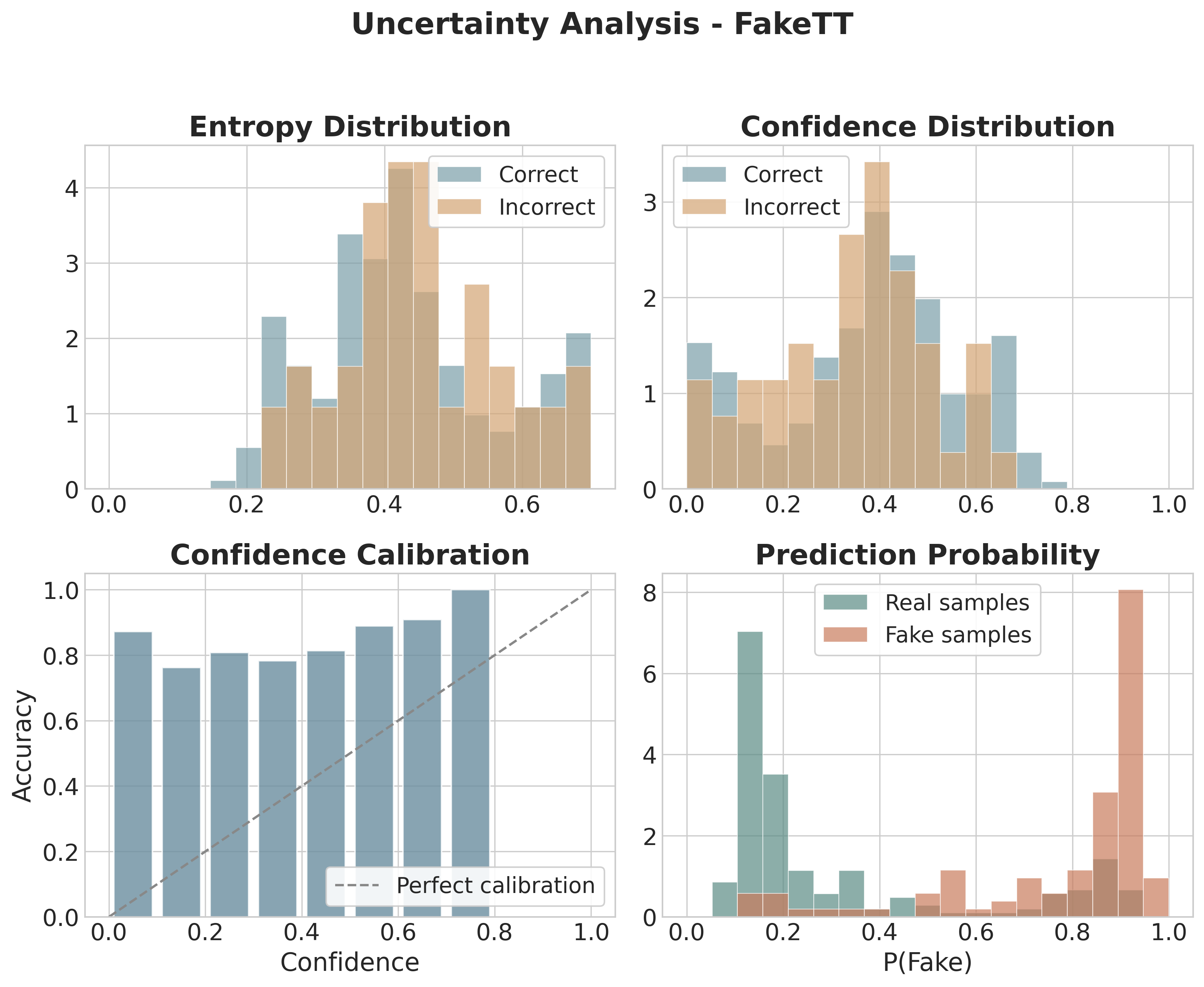}
  \end{subfigure}

  \vspace{0.6em}

  \begin{subfigure}{0.48\textwidth}
    \centering
    \includegraphics[width=\linewidth]{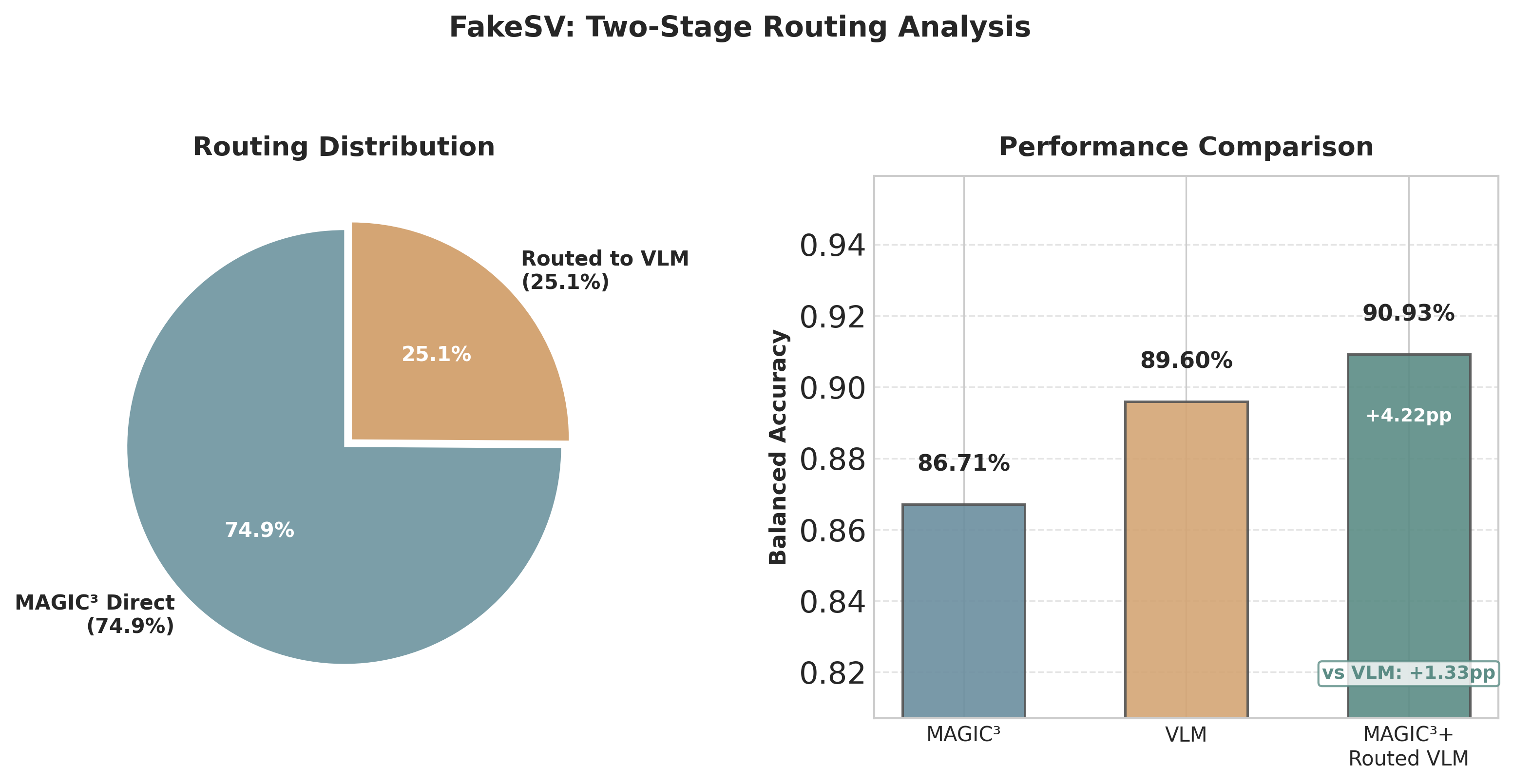}
  \end{subfigure}
  \hfill
  \begin{subfigure}{0.48\textwidth}
    \centering
    \includegraphics[width=\linewidth]{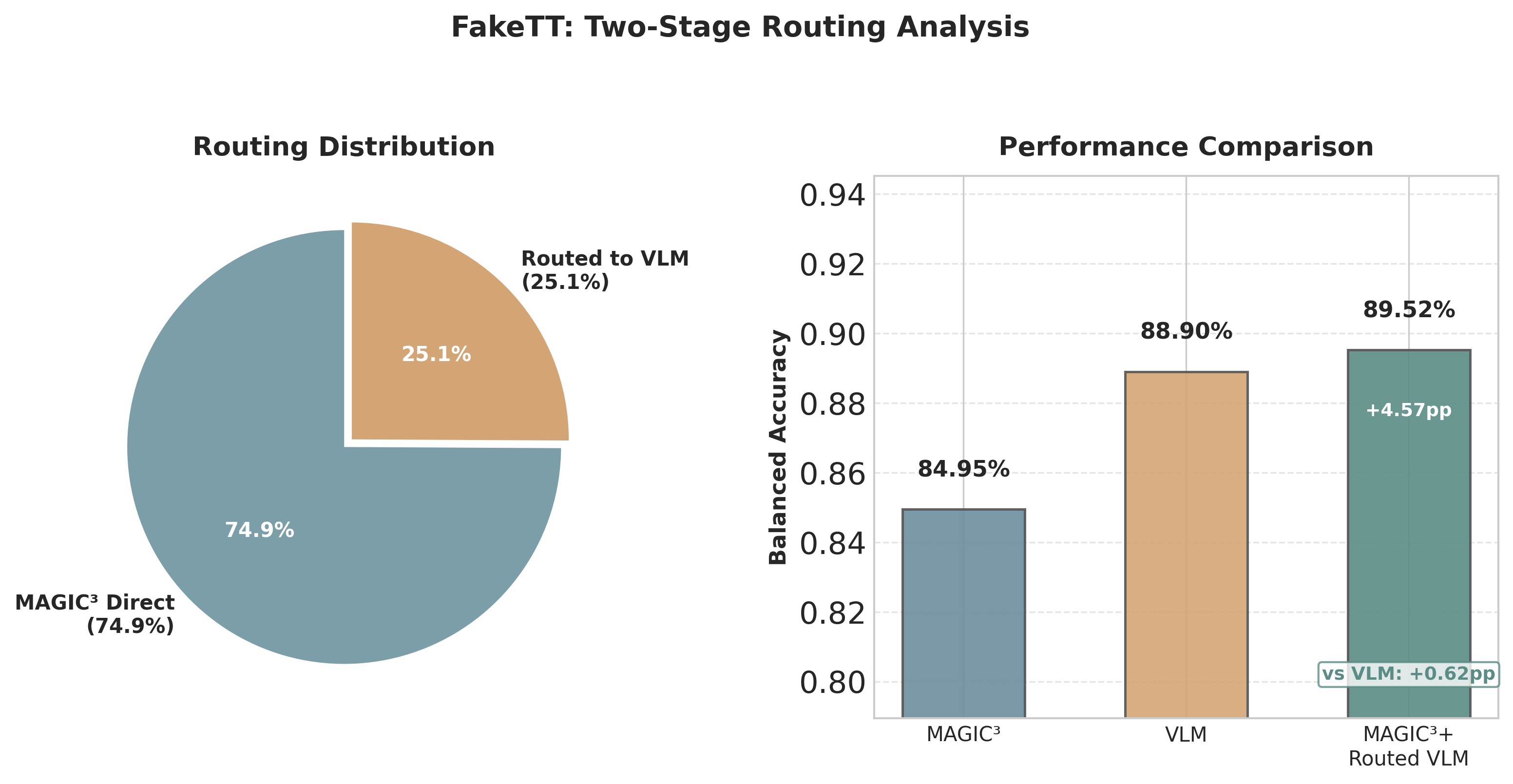}
  \end{subfigure}

  \caption{Uncertainty and two-stage routing behaviour of \model{}. Top: entropy/confidence distributions and calibration plots. Bottom: split between direct predictions and routed samples, and accuracies for routed vs.\ non-routed subsets when using a VLM-based stage-2 detector. Left: FakeSV; right: FakeTT.}
  \label{fig:uncertainty-routing}
\end{figure*}

\begin{figure*}[t]
  \centering
  \begin{subfigure}{0.48\textwidth}
    \centering
    \includegraphics[width=\linewidth]{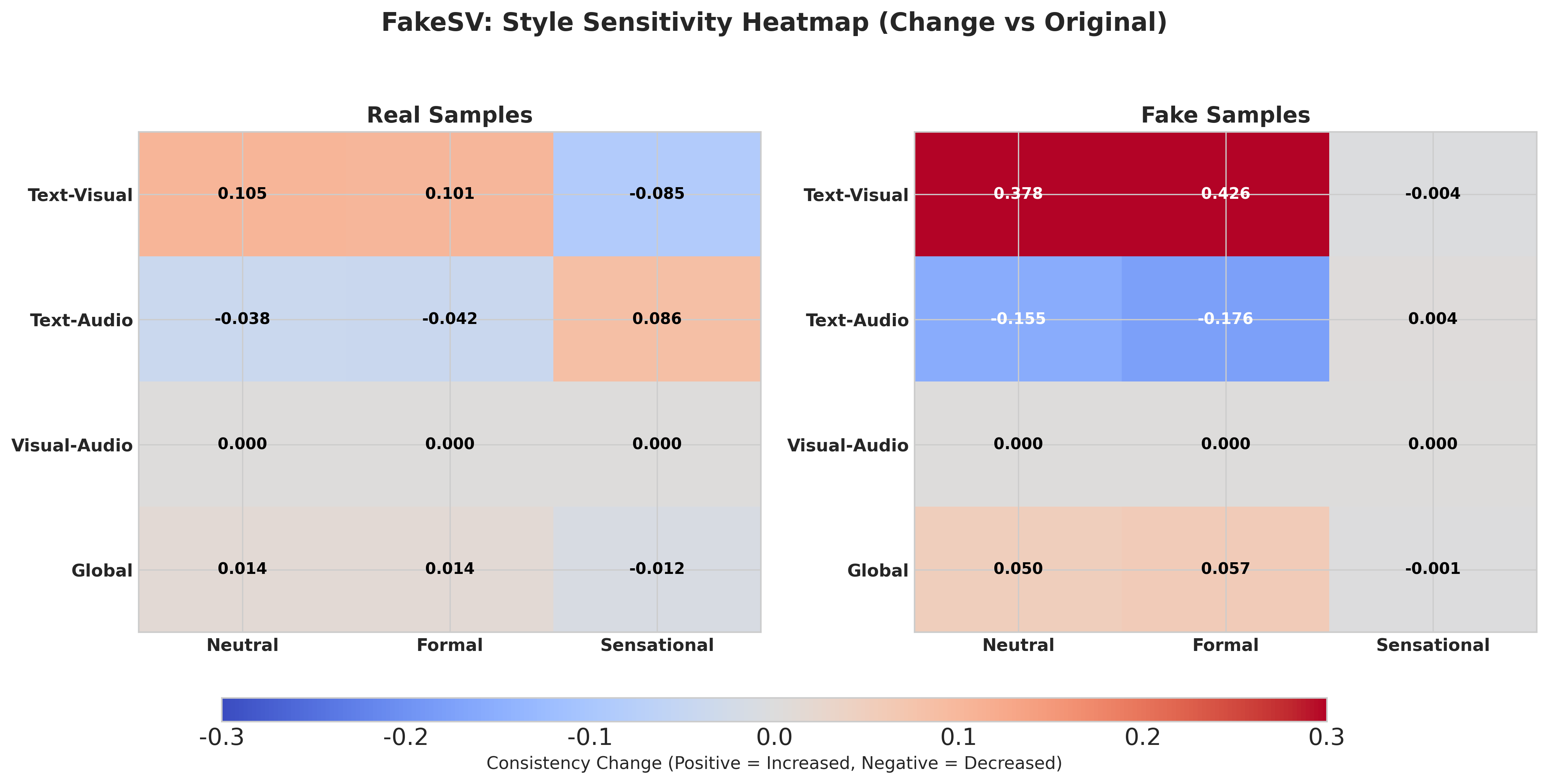}
  \end{subfigure}
  \hfill
  \begin{subfigure}{0.48\textwidth}
    \centering
    \includegraphics[width=\linewidth]{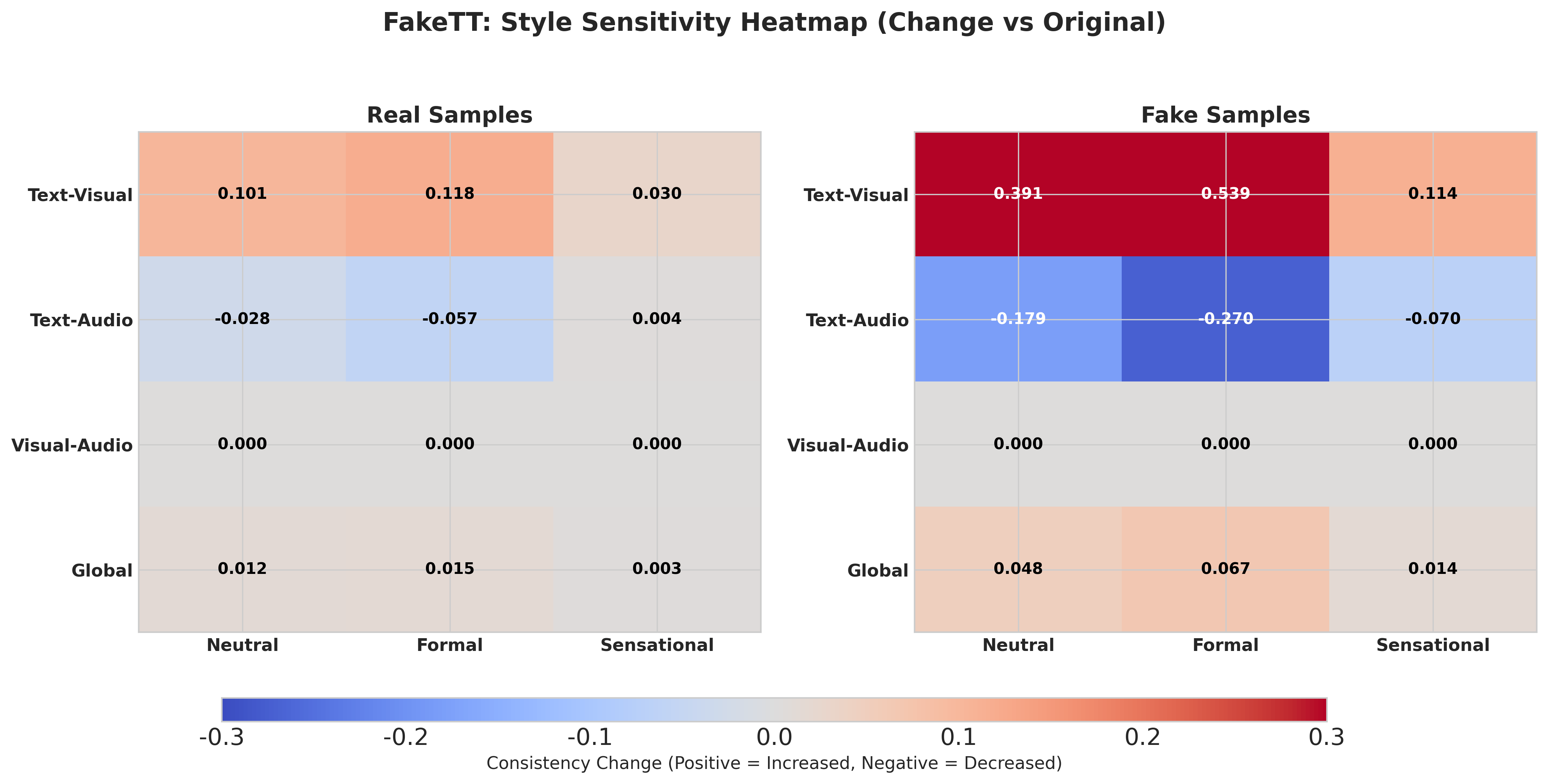}
  \end{subfigure}
  \caption{Style sensitivity heatmaps (Finding~4). Rows: consistency types; columns: style rewrites vs.\ original. Fake samples show larger shifts than real ones.}
  \label{fig:style-heatmap}
\end{figure*}

See Figure~\ref{fig:uncertainty-routing} and Figure~\ref{fig:style-heatmap}.

\begin{figure*}[t]
  \centering
  \begin{subfigure}{0.48\textwidth}
    \centering
    \includegraphics[width=\linewidth]{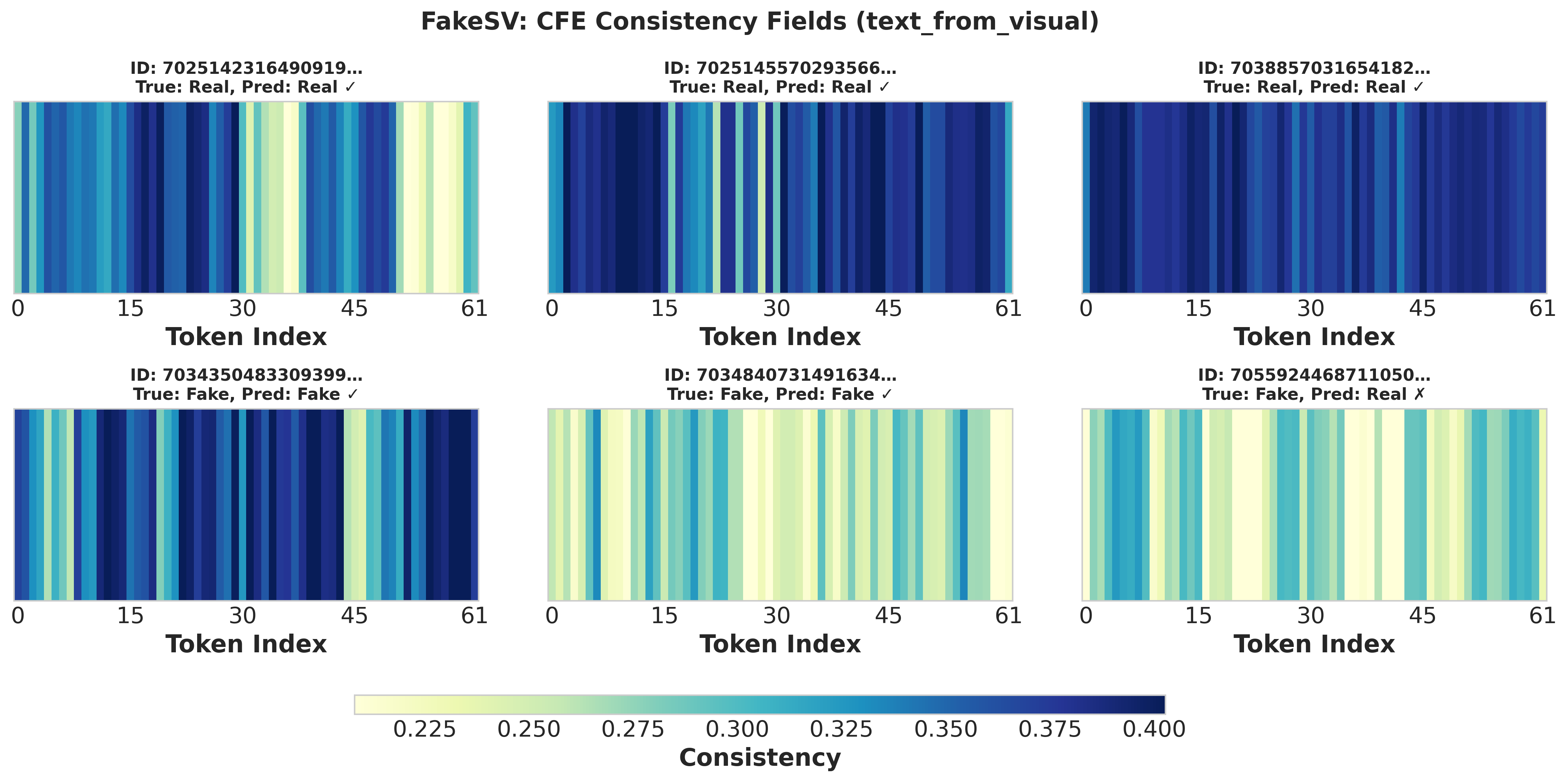}
  \end{subfigure}
  \hfill
  \begin{subfigure}{0.48\textwidth}
    \centering
    \includegraphics[width=\linewidth]{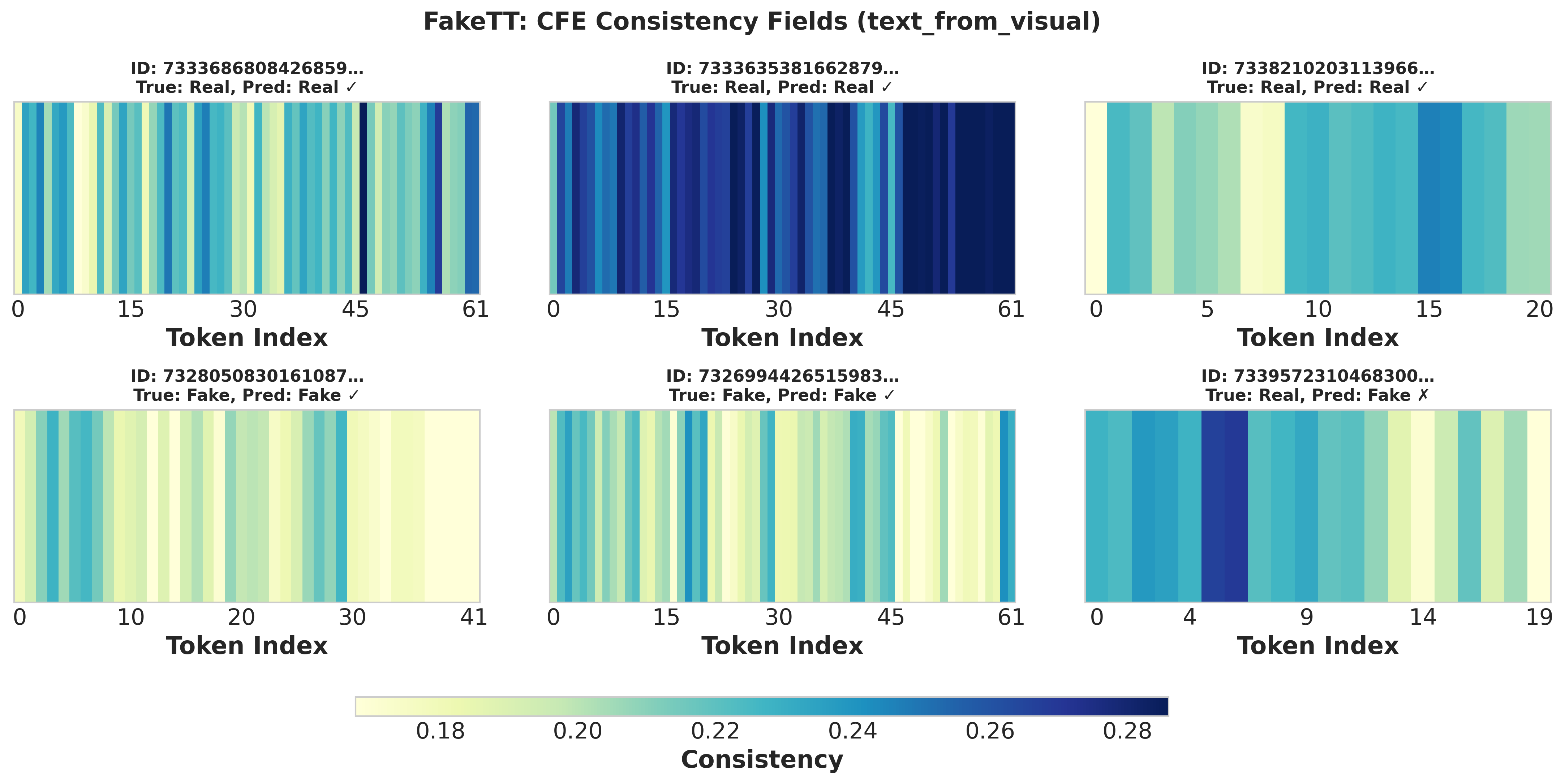}
  \end{subfigure}
  \caption{CFE consistency fields. Colors indicate per-token consistency with the visual stream; sharp high-score regions suggest localized support, while diffuse/low scores indicate potential inconsistencies.}
  \label{fig:cfe-fields}
\end{figure*}

\section{Method Details}
\label{sec:method-details}

This section provides the mathematical details of the components sketched in Section~\ref{sec:methodology}.

\subsection{Cross-Modal Consistency Gate (CMCG)}

We first compute pooled representations for each modality:
\begin{align}
  \bar{\mathbf{h}}_{\mathrm{text}} &= \mathrm{Pool}\!\left(\mathbf{H}_{\mathrm{text}}\right), &
  \bar{\mathbf{h}}_{\mathrm{vis}}  &= \mathrm{Pool}\!\left(\mathbf{H}_{\mathrm{vis}}\right), \nonumber\\
  \bar{\mathbf{h}}_{\mathrm{aud}}  &= \mathrm{Pool}\!\left(\mathbf{H}_{\mathrm{aud}}\right).
\end{align}
Here $\mathrm{Pool}(\cdot)$ denotes mean pooling or a special CLS token.

For each modality pair $(m_i,m_j)\in\{(\mathrm{text},\mathrm{vis}), (\mathrm{text},\mathrm{aud}), (\mathrm{vis},\mathrm{aud})\}$, CMCG computes a scalar consistency score
\begin{equation}
  c_{ij}
  =
  \sigma\!\Bigl(
    \mathbf{w}_{ij}^{\top}
    [\bar{\mathbf{h}}_{m_i};\bar{\mathbf{h}}_{m_j}]
    + b_{ij}
  \Bigr),
\end{equation}
where $[\cdot;\cdot]$ is concatenation and $\sigma$ is the sigmoid function.  
This yields a 3-dimensional consistency vector
\begin{equation}
  \mathbf{c} = [c_{\mathrm{tv}}, c_{\mathrm{ta}}, c_{\mathrm{va}}] \in \mathbb{R}^{3}.
\end{equation}
A small MLP then produces a global consistency score
\begin{equation}
  c_{\mathrm{global}} = \sigma(\mathrm{MLP}_{c}(\mathbf{c})).
\end{equation}

\subsection{Consistency Field Estimator (CFE)}

Suppose we have a cross-attention matrix from modality $m_i$ to $m_j$,
\begin{equation}
  \mathbf{A}_{m_i \leftarrow m_j} \in \mathbb{R}^{L_{m_i} \times L_{m_j}},
\end{equation}
produced by the cross-modal transformer.  
Row $t$ corresponds to a token or frame in modality $m_i$.  
We define the consistency field of $m_i$ with respect to $m_j$ as
\begin{equation}
  F_{ij}^{(m_i)}(t)
  =
  \max_{k}\, \mathbf{A}_{m_i \leftarrow m_j}(t,k),
  \quad t = 1,\dots,L_{m_i}.
\end{equation}
Intuitively, $F_{ij}^{(m_i)}(t)\in[0,1]$ reflects how strongly token $t$ in modality $m_i$ finds supporting evidence in modality $m_j$.  
For training, we add a mild regularizer aligning global CMCG scores with aggregated fields:
\begin{equation}
  \mathcal{L}_{\mathrm{reg}}
  =
  \sum_{(i,j)}
  \bigl(
    c_{ij}
    - \mathrm{mean}(F_{ij}^{(m_i)})
  \bigr)^{2}.
\end{equation}

\subsection{Multi-view Adversarial-Aware Rewrite Fusion (AARF)}

Let $\mathbf{H}_{\mathrm{text}}$ be the original text sequence and $\mathbf{H}_{\mathrm{rew}}^{(v)}$ the $v$-th rewrite.  
We pool them into vectors
\begin{align}
  \mathbf{h}_{\mathrm{orig}} &= \mathrm{Pool}(\mathbf{H}_{\mathrm{text}}), \\
  \mathbf{h}_{\mathrm{rew}}^{(v)} &= \mathrm{Pool}\bigl(\mathbf{H}_{\mathrm{rew}}^{(v)}\bigr).
\end{align}
For each view $v$, we estimate a scalar rewrite quality
\begin{equation}
  q^{(v)}
  =
  \sigma\!\bigl(
    \mathrm{MLP}_{q}
    [\mathbf{h}_{\mathrm{orig}};\mathbf{h}_{\mathrm{rew}}^{(v)}]
  \bigr),
\end{equation}
and compute gating logits
\begin{equation}
  \boldsymbol{g}
  =
  \mathrm{MLP}_{g}\bigl(
    [\mathbf{h}_{\mathrm{orig}};\mathbf{h}_{\mathrm{rew}}^{(1)};\dots;
     \mathbf{h}_{\mathrm{rew}}^{(V)};q^{(1)},\dots,q^{(V)}]
  \bigr).
\end{equation}
We obtain preliminary weights $\boldsymbol{\alpha}^{\mathrm{soft}} = \mathrm{softmax}(\boldsymbol{g})$, enforce a minimum weight $\alpha_{\min}$ on the original text, and renormalize:
\begin{align}
  \tilde{\alpha}_0 &= \max(\alpha_0^{\mathrm{soft}}, \alpha_{\min}),\\
  \tilde{\alpha}_v &= \alpha_v^{\mathrm{soft}}, \quad v\ge1,\\
  \alpha_v &=
  \begin{cases}
    \tilde{\alpha}_0, & v=0, \\
    (1-\tilde{\alpha}_0)\,
    \dfrac{\tilde{\alpha}_v}{\sum_{u=1}^{V} \tilde{\alpha}_u},
    & v\ge 1.
  \end{cases}
\end{align}
The fused representation is
\begin{equation}
  \mathbf{h}_{\mathrm{text}}^{\mathrm{fuse}}
  =
  \alpha_{0}\,\mathbf{h}_{\mathrm{orig}}
  +
  \sum_{v=1}^{V}
    \alpha_{v}\,\mathrm{Proj}\bigl(\mathbf{h}_{\mathrm{rew}}^{(v)}\bigr).
\end{equation}
An InfoNCE-style loss encourages all views of the same video to be close and views of different videos to be separated.

\subsection{Hierarchical Multimodal Transformer (HMT)}

Layer A performs intra-modal refinement:
\begin{align}
  \mathbf{H}_{\mathrm{text}}^{(A)} &= \mathrm{TransEnc}_{\mathrm{text}}(\mathbf{H}_{\mathrm{text}}), \\
  \mathbf{H}_{\mathrm{vis}}^{(A)}  &= \mathrm{TransEnc}_{\mathrm{vis}}(\mathbf{H}_{\mathrm{vis}}), \\
  \mathbf{H}_{\mathrm{aud}}^{(A)}  &= \mathrm{TransEnc}_{\mathrm{aud}}(\mathbf{H}_{\mathrm{aud}}).
\end{align}
Layer B performs consistency-weighted cross-modal attention; for modalities $m_i,m_j$ we compute
\begin{equation}
  \tilde{\mathbf{H}}_{m_i \leftarrow m_j}
  = \mathrm{CrossAttn}\bigl(
      \mathbf{H}_{m_i}^{(A)},
      \mathbf{H}_{m_j}^{(A)}
    \bigr).
\end{equation}
Attention weights are stored for CFE.  
We then aggregate messages using weights derived from $\mathbf{c}$:
\begin{align}
  \boldsymbol{\beta}_{m} &= \mathrm{softmax}\bigl(\mathrm{MLP}_{m}(\mathbf{c})\bigr), \\
  \mathbf{H}_{m}^{(B)}   &= \sum_{j\neq m} \beta_{mj}\,\tilde{\mathbf{H}}_{m \leftarrow j}.
\end{align}

Layer C pools refined representations and fuses them with the AARF text:
\begin{align}
\mathbf{Z}
  &= \begin{bmatrix}
        \mathrm{Pool}(\mathbf{H}_{\mathrm{text}}^{(B)}) \\
        \mathrm{Pool}(\mathbf{H}_{\mathrm{vis}}^{(B)}) \\
        \mathrm{Pool}(\mathbf{H}_{\mathrm{aud}}^{(B)}) \\
        \mathbf{h}_{\mathrm{text}}^{\mathrm{fuse}}
     \end{bmatrix},\\
\mathbf{h}_{\mathrm{global}}
  &= \mathrm{TransEnc}_{\mathrm{global}}(\mathbf{Z})_{\mathrm{[CLS]}}.
\end{align}

\subsection{Contrastive--Adversarial Joint Learning (CAJL)}

Let $y_i\in\{0,1\}$ be labels and $\hat{y}_i$ predicted probabilities.  
The cross-entropy loss is
\begin{equation}
  \mathcal{L}_{\mathrm{ce}}
  =
  -\frac{1}{N}
  \sum_{i}
  \Bigl(
    y_i \log \hat{y}_i
    + (1-y_i)\log (1-\hat{y}_i)
  \Bigr).
\end{equation}
We use InfoNCE for contrastive terms.  
For representation pairs $(\mathbf{u}_i,\mathbf{v}_i)$ from the same video and negatives from other videos,
\begin{equation}
  \mathcal{L}_{\mathrm{NCE}}
  =
  -\frac{1}{N}
  \sum_{i}
  \log
  \frac{\exp(\mathrm{sim}(\mathbf{u}_i,\mathbf{v}_i) / \tau)}
  {\sum_j \exp(\mathrm{sim}(\mathbf{u}_i,\mathbf{v}_j)/\tau)}.
\end{equation}
$\mathcal{L}_{\mathrm{intra}}$ uses different views of the same video (original text, rewrites, audio, visuals), and $\mathcal{L}_{\mathrm{cross}}$ aligns fused text and global representations.

For adversarial consistency, we apply small perturbations $\delta_i$ to intermediate representations and enforce prediction consistency:
\begin{equation}
  \mathcal{L}_{\mathrm{adv}}
  =
  \frac{1}{N}
  \sum_i
  \mathrm{KL}\bigl(
    p_\theta(y\mid\mathbf{h}_i)
    \Vert
    p_\theta(y\mid\mathbf{h}_i+\delta_i)
  \bigr).
\end{equation}
We also encourage consistent predictions between original and fused text:
\begin{equation}
  \mathcal{L}_{\mathrm{sem}}
  =
  \frac{1}{N}
  \sum_i
  \mathrm{KL}\bigl(
    p_\theta(y\mid\mathbf{h}_{\mathrm{orig},i})
    \Vert
    p_\theta(y\mid\mathbf{h}_{\mathrm{text},i}^{\mathrm{fuse}})
  \bigr).
\end{equation}
A style invariance term applies InfoNCE over projections of all text views:
\begin{equation}
  \mathcal{L}_{\mathrm{style}}
  =
  \frac{1}{N}
  \sum_i
  \mathcal{L}_{\mathrm{NCE}}\bigl(\{\mathbf{z}_{\mathrm{text},i}^{(v)}\}_{v=0}^{V}\bigr).
\end{equation}

The final loss is
\begin{align}
  \mathcal{L}
  =
  \mathcal{L}_{\mathrm{ce}}
  +
  \lambda_{\mathrm{intra}} \mathcal{L}_{\mathrm{intra}}
  +
  \lambda_{\mathrm{cross}} \mathcal{L}_{\mathrm{cross}}
  +
  \lambda_{\mathrm{adv}} \mathcal{L}_{\mathrm{adv}} \nonumber\\
  +
  \lambda_{\mathrm{sem}} \mathcal{L}_{\mathrm{sem}}
  +
  \lambda_{\mathrm{style}} \mathcal{L}_{\mathrm{style}}
  +
  \lambda_{\mathrm{reg}} \mathcal{L}_{\mathrm{reg}}.
\end{align}
Hyperparameters are given in Table~\ref{tab:hyperparams}.

\section{Additional Quantitative Analyses for Findings}
\label{sec:addl-findings-stats}

This appendix reports the quantitative statistics that support Findings~1--4.

\subsection{Aggregate Consistency Statistics}
\label{subsec:agg-stats}

Table~\ref{tab:finding1-stats} lists mean $\pm$ standard deviation of the four consistency scores for real vs.\ fake videos, along with two-sample $t$-tests. The text--visual vs.\ text--audio asymmetry is highly significant on both datasets; visual--audio is nearly identical across classes.

\begin{table}[h]
  \centering
  \footnotesize
  \setlength{\tabcolsep}{3pt}
  \renewcommand{\arraystretch}{0.9}
  \begin{tabular}{llrrr}
    \toprule
    Dataset & Score & Real & Fake & $p$-value \\
    \midrule
    \multirow{4}{*}{FakeSV}
      & $c_{\mathrm{tv}}$     & $0.82\!\pm\!0.32$ & $0.16\!\pm\!0.27$ & $p \ll 0.001$ \\
      & $c_{\mathrm{ta}}$     & $0.50\!\pm\!0.20$ & $0.88\!\pm\!0.13$ & $p \ll 0.001$ \\
      & $c_{\mathrm{va}}$     & $0.89\!\pm\!0.02$ & $0.89\!\pm\!0.02$ & $p=0.82$ \\
      & $c_{\mathrm{global}}$ & $0.47\!\pm\!0.04$ & $0.38\!\pm\!0.04$ & $p \ll 0.001$ \\
    \midrule
    \multirow{4}{*}{FakeTT}
      & $c_{\mathrm{tv}}$     & $0.84\!\pm\!0.29$ & $0.31\!\pm\!0.35$ & $p \ll 0.001$ \\
      & $c_{\mathrm{ta}}$     & $0.39\!\pm\!0.23$ & $0.77\!\pm\!0.18$ & $p \ll 0.001$ \\
      & $c_{\mathrm{va}}$     & $0.86\!\pm\!0.03$ & $0.86\!\pm\!0.03$ & $p=0.49$ \\
      & $c_{\mathrm{global}}$ & $0.47\!\pm\!0.04$ & $0.40\!\pm\!0.04$ & $p \ll 0.001$ \\
    \bottomrule
  \end{tabular}
  \caption{Finding~1 statistics: consistency scores by class (mean$\pm$std) and two-sample $t$-test $p$-values. Text--visual vs.\ text--audio patterns flip between real and fake videos on both datasets.}
  \label{tab:finding1-stats}
\end{table}

\subsection{Correlation Between Global Consistency and Predictions}
\label{subsec:cglobal-corr}

Table~\ref{tab:finding2-corr} reports Spearman/Pearson correlations between $c_{\mathrm{global}}$ and predicted fake probability, along with a linear-fit $R^2$. Table~\ref{tab:finding2-bins} shows the empirical fake rate in five $c_{\mathrm{global}}$ bins, illustrating the generally decreasing trend.

\begin{table}[h]
  \centering
  \footnotesize
  \setlength{\tabcolsep}{3pt}
  \renewcommand{\arraystretch}{0.9}
  \begin{tabular}{lrrr}
    \toprule
    Dataset & Spear. $\rho$ & Pear. $r$ & $R^2$ \\
    \midrule
    FakeSV & $-0.91$ ($p \ll 0.001$) & $-0.99$ ($p \ll 0.001$) & $0.99$ \\
    FakeTT & $-0.96$ ($p \ll 0.001$) & $-0.96$ ($p \ll 0.001$) & $0.91$ \\
    \bottomrule
  \end{tabular}
  \caption{Finding~2: correlation between $c_{\mathrm{global}}$ and predicted fake probability.}
  \label{tab:finding2-corr}
\end{table}

\begin{table}[h]
  \centering
  \footnotesize
  \setlength{\tabcolsep}{3pt}
  \renewcommand{\arraystretch}{0.9}
  \begin{tabular}{lrr}
    \toprule
    $c_{\mathrm{global}}$ bin & FakeSV rate & FakeTT rate \\
    \midrule
    Lowest (Q1) & 0.93 & 0.83 \\
    Q2          & 0.89 & 0.55 \\
    Q3          & 0.76 & 0.17 \\
    Q4          & 0.19 & 0.03 \\
    Highest (Q5) & 0.05 & 0.07 \\
    \bottomrule
  \end{tabular}
  \caption{Finding~2: fake rates in five quantile bins of $c_{\mathrm{global}}$ (lower scores = more inconsistent).}
  \label{tab:finding2-bins}
\end{table}

\subsection{Style Sensitivity Statistics}
\label{subsec:style-stats}

Table~\ref{tab:finding4-style} compares style-induced variance (across original+three rewrites) for real vs.\ fake videos. Fake samples show consistently higher variance for text--visual, text--audio, and global consistency on both datasets; visual--audio is unchanged, as expected.

\begin{table}[h]
  \centering
  \footnotesize
  \setlength{\tabcolsep}{3pt}
  \renewcommand{\arraystretch}{0.9}
  \begin{tabular}{llrrr}
    \toprule
    Dataset & Score & Real var & Fake var & $p$ \\
    \midrule
    \multirow{4}{*}{FakeSV}
      & $c_{\mathrm{tv}}$     & 0.0336 & 0.0815 & $p<10^{-16}$ \\
      & $c_{\mathrm{ta}}$     & 0.0109 & 0.0148 & $p<10^{-2}$ \\
      & $c_{\mathrm{va}}$     & 0.0    & 0.0    & n.s. \\
      & $c_{\mathrm{global}}$ & $6.19{\times}10^{-4}$ & $1.44{\times}10^{-3}$ & $p<10^{-15}$ \\
    \midrule
    \multirow{4}{*}{FakeTT}
      & $c_{\mathrm{tv}}$     & 0.0194 & 0.0801 & $p<10^{-21}$ \\
      & $c_{\mathrm{ta}}$     & 0.0081 & 0.0170 & $p<10^{-7}$ \\
      & $c_{\mathrm{va}}$     & 0.0    & 0.0    & n.s. \\
      & $c_{\mathrm{global}}$ & $3.13{\times}10^{-4}$ & $1.23{\times}10^{-3}$ & $p<10^{-20}$ \\
    \bottomrule
  \end{tabular}
  \caption{Finding~4: variance of consistency scores under style rewrites (real vs.\ fake). ``n.s.'' denotes non-significant differences ($p \ge 0.05$).}
  \label{tab:finding4-style}
\end{table}

\subsection{Two-Stage Routing Analysis Details}
\label{subsec:finding3-routing}

We treat the routing threshold as a hyperparameter and tune it on the validation set to route approximately 25\% of samples on both datasets. Table~\ref{tab:finding3-routing} summarizes the routing strategies used. For FakeSV, we use entropy-based routing (high entropy indicates uncertain predictions); for FakeTT, we use a \emph{difficulty score} that combines entropy, global consistency, and confidence: $d = \text{entropy}_{\text{norm}} + (1 - c_{\text{global},\text{norm}}) + (1 - \text{conf}_{\text{norm}})$, routing samples with the highest difficulty scores.

\begin{table}[h]
  \centering
  \footnotesize
  \setlength{\tabcolsep}{4pt}
  \renewcommand{\arraystretch}{0.9}
  \resizebox{\columnwidth}{!}{%
    \begin{tabular}{lrrl}
      \toprule
      Dataset & Routed Ratio & \model{}+VLM Acc & Routing Strategy \\
      \midrule
      FakeSV & 25.1\% & 90.93\% & Entropy-based \\
      FakeTT & 25.1\% & 89.52\% & Difficulty score \\
      \bottomrule
    \end{tabular}
  }
  \caption{Finding~3: routing strategies and routed ratios used to construct the two-stage systems. Both strategies outperform the standalone VLM detector (FakeSV: +1.33pp, FakeTT: +0.62pp).}
  \label{tab:finding3-routing}
\end{table}

\section{Datasets and Splits}
\label{sec:datasets}

Table~\ref{tab:datasets} summarizes the two benchmarks used in this paper.  
We follow the official chronological splits from~\citet{huang2025knowledge}, using 70\%/15\%/15\% for training, validation, and testing.

\begin{table}[h]
  \centering
  \small
  \begin{tabular}{lcc}
    \toprule
    \textbf{Statistic} & \textbf{FakeSV} & \textbf{FakeTT} \\
    \midrule
    Time range & 2017/10--2022/02 & 2019/05--2024/03 \\
    Avg. duration (s) & 39.88 & 47.69 \\
    \# Fake videos & 1{,}810 & 1{,}172 \\
    \# Real videos & 1{,}814 & 819 \\
    \# Total & 3{,}624 & 1{,}991 \\
    \bottomrule
  \end{tabular}
  \caption{Statistics of FakeSV and FakeTT. All videos are short-form clips on real-world platforms.}
  \label{tab:datasets}
\end{table}

\section{LLM Rewriting Prompts}
\label{sec:prompt}

We use three prompts to obtain multi-view rewrites. In our experiments we use DeepSeek-V3.2; for FakeSV we use Chinese translations of these prompts.

\paragraph{Neutral news style.}
\begin{quote}
\small
You are a professional news editor.\\
Rewrite the following video description in a neutral, objective news style.\\
Focus on clarity and factual accuracy. Do not add new information.\\[0.3em]
Original text: \{input\_text\}\\[0.3em]
Rewritten text:
\end{quote}

\paragraph{Formal elaborated style.}
\begin{quote}
\small
You are a senior journalist.\\
Rewrite the following video description in a more formal and elaborated style,\\
making the key facts explicit and clearly structured. Avoid emotional language.\\[0.3em]
Original text: \{input\_text\}\\[0.3em]
Rewritten text:
\end{quote}

\paragraph{Sensational style.}
\begin{quote}
\small
You are writing a sensational social media post about breaking news.\\
Rewrite the following video description to sound more emotional and attention-grabbing,\\
while keeping the core facts unchanged.\\[0.3em]
Original text: \{input\_text\}\\[0.3em]
Rewritten text:
\end{quote}

\section{Implementation Details and Hyperparameters}
\label{sec:impl-details}

Backbone encoders use \texttt{bert-base-multilingual-uncased} for text (maximum length 256), Swin Transformer~\citep{liu2021swin} for $K$ keyframes per video, and ExHuBERT-based emotion embeddings for audio~\citep{amiriparian2024exhubert}.  
These encoders are used only for offline feature extraction; during \model{} training we operate purely on the pre-extracted features and do not update encoder parameters.

Training uses AdamW with learning rate $5\times10^{-5}$, batch size 32, cosine decay with 10\% warm-up, dropout 0.3 in transformer layers, and early stopping based on validation macro-F1.  
We train for up to 50 epochs and keep the checkpoint with best validation macro-F1.  
In our main experiments we set $\lambda_{\mathrm{intra}} = \lambda_{\mathrm{cross}} = 0.1$, $\lambda_{\mathrm{adv}} = 0.5$, $\lambda_{\mathrm{sem}} = 0.1$, $\lambda_{\mathrm{style}} = 0.1$, and $\lambda_{\mathrm{reg}} = 0.05$, with $\alpha_{\min}=0.5$, $V=3$, and $K_{\mathrm{MC}}=5$.

\begin{table}[h]
  \centering
  \small
  \begin{tabular}{l|c}
    \toprule
    \textbf{Hyperparameter} & \textbf{Value} \\
    \midrule
    Learning rate & $5 \times 10^{-5}$ \\
    Batch size & 32 \\
    Epochs & 50 \\
    Warmup ratio & 0.1 \\
    Hidden dimension & 256 \\
    Attention heads (HMT) & 8 \\
    Dropout & 0.3 \\
    $\lambda_{\mathrm{intra}}$ & 0.1 \\
    $\lambda_{\mathrm{cross}}$ & 0.1 \\
    $\lambda_{\mathrm{adv}}$ & 0.5 \\
    $\lambda_{\mathrm{sem}}$ & 0.1 \\
    $\lambda_{\mathrm{style}}$ & 0.1 \\
    $\lambda_{\mathrm{reg}}$ & 0.05 \\
    $\alpha_{\min}$ (AARF) & 0.5 \\
    $V$ (rewrites per video) & 3 \\
    \bottomrule
  \end{tabular}
  \caption{Key hyperparameter settings for \model{}.}
  \label{tab:hyperparams}
\end{table}

\section{Efficiency and Deployment Discussion}
\label{sec:efficiency}
Because all encoders are used only for offline feature extraction and \model{} operates on pre-extracted features, inference on cached features reduces to a single detector head.  
The trainable detector head has roughly 185M parameters (with no encoder fine-tuning), which is lightweight compared to multi-billion-parameter VLM backbones. On a single RTX 5090 GPU, a full training run on each dataset (including validation and test evaluation) takes about 15 minutes. 
With inference batch size 128 on the same GPU, we measure about 680 samples/s and $\sim$1.5\,ms per sample, with $\sim$1.67\,GB peak GPU memory on cached features.  
Peak GPU memory usage during training is about 11\,GB on the same GPU.  
End-to-end latency is dominated by video decoding and feature extraction, which can be amortized via caching and asynchronous preprocessing in practical systems.

In contrast, VLM-based detectors typically require end-to-end processing with a large vision--language backbone, which is more expensive to train (often requiring multi-GPU setups) and to serve at scale.  
Operating purely at the feature level allows \model{} to be retrained quickly for new topics or sub-verticals within the short-video misinformation space, making it a practical choice for platforms that need rapid adaptation under hardware constraints.
\end{document}